\documentclass[lettersize,journal]{IEEEtran}
\usepackage{amsmath,amsfonts}
\usepackage{algorithmic}
\usepackage{algorithm}
\usepackage{array}
\usepackage{textcomp}
\usepackage{stfloats}
\usepackage{url}
\usepackage{verbatim}
\usepackage{graphicx}
\usepackage[labelformat=simple]{subcaption}

\usepackage{cite}
\graphicspath{ {images/} }
\usepackage{tikz}
\usetikzlibrary{arrows,positioning,fit,shapes.geometric}
\usetikzlibrary{shapes}
\tikzset{>=latex}

\begin{document}
	
\title{Exploring Generative Adversarial Networks for Text-to-Image Generation with Evolution Strategies}

\author{Victor~Costa,~\IEEEmembership{}%
	Nuno~Louren\c{c}o,~\IEEEmembership{}%
	Jo\~{a}o Correia,~\IEEEmembership{}%
	and~Penousal~Machado~\IEEEmembership{}
	\thanks{This work is partially funded by the project grant DSAIPA/DS/0022/2018 (GADgET), by national funds through the FCT - Foundation for Science and Technology, I.P., within the scope of the project CISUC - UID/CEC/00326/2020 and by European Social Fund, through the Regional Operational Program Centro 2020. We also thank the NVIDIA Corporation for the hardware granted to this research.}%
	\thanks{V. Costa, N. Louren\c{c}o, and P. Machado are with University of Coimbra, Centre for Informatics and Systems of the University of Coimbra, Department of Informatics Engineering, Coimbra, Portugal (e-mail: \{vfc, naml, jncor, machado\}@dei.uc.pt).}
}



\maketitle

\begin{abstract}
In the context of generative models, text-to-image generation achieved impressive results in recent years.
Models using different approaches were proposed and trained in huge datasets of pairs of texts and images.
However, some methods rely on pre-trained models such as Generative Adversarial Networks, searching through the latent space of the generative model by using a gradient-based approach to update the latent vector, relying on loss functions such as the cosine similarity.
In this work, we follow a different direction by proposing the use of Covariance Matrix Adaptation Evolution Strategy to explore the latent space of Generative Adversarial Networks.
We compare this approach to the one using Adam and a hybrid strategy.
We design an experimental study to compare the three approaches using different text inputs for image generation by adapting an evaluation method based on the projection of the resulting samples into a two-dimensional grid to inspect the diversity of the distributions.
The results evidence that the evolutionary method achieves more diversity in the generation of samples, exploring different regions of the resulting grids.
Besides, we show that the hybrid method combines the explored areas of the gradient-based and evolutionary approaches, leveraging the quality of the results.
\end{abstract}


\section{Introduction}
\IEEEPARstart{R}{ecently}, generative models gained a lot of attention from the community, mainly in the image domain.
In this context, several models were proposed that are capable of generating images based on a text input~\cite{ramesh2021zero,ramesh2022hierarchical}.
Some rely on pre-trained generative models, such as Generative Adversarial Networks (GANs), combined with a language processing system capable of describing an input image using natural language.
For this, a mapping mechanism from the output of the language model to the latent space of a GAN is paramount for achieving a meaningful representation of the resulting images.

A recent proposal presented on~\cite{bigsleep} uses Contrastive Language-Image Pre-training (CLIP)~\cite{radford2021learning} and BigGAN~\cite{brock2018large} to convert texts to images.
In this model, the latent space of the GAN is the subject of training using backpropagation, namely the Adam optimizer~\cite{kingma2015adam}, according to the features extracted by CLIP given a sentence.

In this work, we propose the use of Covariance Matrix Adaptation Evolution Strategy (CMA-ES)~\cite{hansen2001completely} instead of Adam to explore the latent space of GANs when building a system for text-to-image generation, training the model to achieve a diverse representation for the text input.
For this, we adapt the model proposed in~\cite{bigsleep} by modifying the latent inputs provided to BigGAN to enable the use of the gradient-based approach and also CMA-ES.
We also propose a hybrid approach combining the two strategies.

We conduct an experimental study comparing the evolutionary approach, the regular backpropagation training, and the hybrid solution.
To represent our results, we adapt the evaluation method proposed in~\cite{costa2021tsne}.
This evaluation method relies on the t-Distributed Stochastic Neighbour Embedding (t-SNE)~\cite{maaten2008visualizing} algorithm for projecting the generated samples into a two-dimensional grid.
In this way, we provide insight into the distribution achieved by the methods evaluated in our study not only by visual inspection but also by a quantitative metric.

The experiments evaluated three approaches for mapping the latent space: CMA-ES, Adam, and a Hybrid model.
The CMA-ES approach evolves a population of individuals representing the latent space of the text input.
The second approach uses the Adam optimizer to discover the best representation of the latent space for the text input.
Finally, the hybrid approach combines CMA-ES with Adam on the exploration of the latent space.
We use the image created by the final iteration of every execution for each approach, creating the two-dimensional projection of the images using the evaluation method adapted from~\cite{costa2021tsne}.

The results show that Adam is limited in the exploration of the latent space when compared with the evolutionary method.
The CMA-ES model is able to achieve more diversity in the results, leading to a better representation of the input sentence.
The hybrid model also achieved better diversity when compared to the Adam approach but displayed better quality than the CMA-ES model.

\section{Background and Related Work}
Generative models achieved impressive results in the last years with the introduction of Generative Adversarial Networks (GANs)~\cite{NIPS2014_5423}, Variational AutoEncoders (VAE)~\cite{kingma2013auto}, and recently with the rise of diffusion models~\cite{sohl2015deep,song2019generative,ho2022cascaded}.
Concerning the image domain, generative models such as BigGAN~\cite{brock2018large}, VQGAN~\cite{esser2021taming}, DALL-E~\cite{ramesh2021zero}, and DALL-E 2~\cite{ramesh2022hierarchical} are capable of synthesize high-quality data.

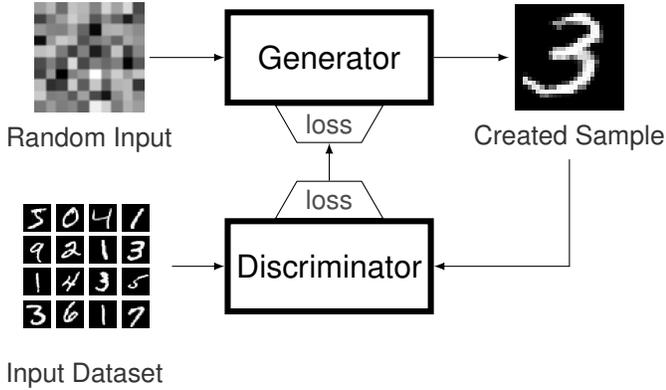
\begin{figure}
	\centering
	\begin{tikzpicture}[
	node distance=8mm,
	title/.style={font=\fontsize{10}{10}\color{black!80}\sffamily},
	block/.style= {draw, rectangle, align=center,minimum width=2.7cm,minimum height=1.2cm,font=\fontsize{12}{12}\sffamily},
	loss/.style= {draw, trapezium, align=center,minimum width=1cm,minimum height=0.3cm,font=\fontsize{10}{10}\color{black!70}\sffamily},
	image/.style= {align=center,minimum width=1cm,minimum height=1cm,inner sep=2pt},
	]
	\node [image] (input) {\includegraphics[width=.08\textwidth]{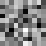}};
	\node [title, below =0cm of input] {Random Input};
	\node [block, right =1.0cm of input, line width=0.8mm]  (generator) {Generator};
	\node [loss, below =-0.01cm of generator, shape border rotate=180]  (lossg) {loss};
	\node [image, right =1.0cm of generator]  (sample) {\includegraphics[width=.08\textwidth]{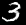}};
	\node [title, below =0cm of sample] (sample-text) {Created Sample};
	\node [block, below =1.0cm of lossg, line width=0.8mm]  (discriminator) {Discriminator};
	\node [loss, above =-0.01cm of discriminator]  (lossd) {loss};
	\node [image, left =0.7cm of discriminator] (dataset) {\includegraphics[width=.12\textwidth]{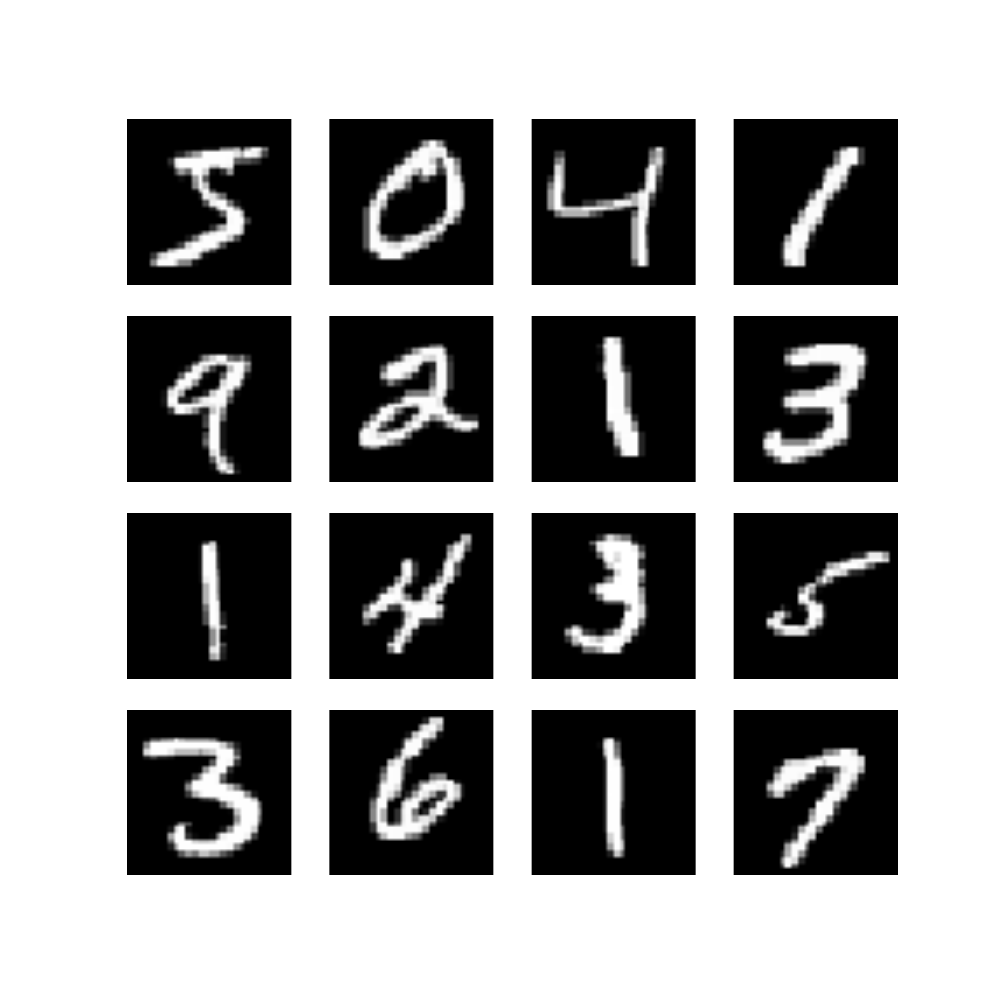}};
	\node [title, below =0cm of dataset] {Input Dataset};
	
	\path[draw,->]
	(input) edge (generator)
	(generator) edge (sample)
	(sample-text) |- (discriminator)
	(dataset) edge (discriminator)
	(lossd) edge (lossg)
	;
	\end{tikzpicture}
	\caption{High-level interaction between the components of a GAN during the training process.} \label{fig:gan}
\end{figure}

The GAN model originally proposed in~\cite{NIPS2014_5423}, uses two neural networks, a generator and a discriminator, trained in an adversarial scenario aiming to capture the distribution of an input dataset.
To achieve this goal, the discriminator is trained to distinguish between data from the input dataset and synthetic samples produced by the generator.
On the other hand, the generator receives a probability distribution (e.g., a normal distribution) as input and uses its neural network to transform it into samples resembling the input dataset.
Figure \ref{fig:gan} depicts the high-level interaction between the components in a GAN when trained using a digits input dataset, such as the MNIST dataset~\cite{lecun1998mnist}.

Over the years, several contributions have emerged to improve the original GAN model.
Conditional GAN~\cite{mirza2014conditional} introduces a mechanism to condition the output sample created by the generator, using an additional input to represent the type of sample that should be created.
BigGAN~\cite{brock2018large} improved the model by employing new techniques to achieve good results in large-scale training, such as the use of orthogonal regularization.
VQGAN~\cite{esser2021taming} makes use of a transformer to enhance the GAN model and produce high-resolution images with impressive quality.

Another example of generative models is the Variational AutoEncoder (VAE).
A VAE is an artificial neural network composed of two parts: an encoder and a decoder.
The goal is to learn the latent space for generating new data based on the input dataset.
In VAE, the input is transformed into probabilistic distributions, and the communication of the encoder and the decoder occurs through a sampling function for regularization in the training process.
DALL-E~\cite{ramesh2021zero} makes use of a VAE to learn a representation of the input dataset, compressing images from 256x256 into a tokenized grid of 32x32.
Then, a transformer is used to learn to correspond the text input and image tokens, producing a powerful text-to-image generative model.

Diffusion models consist of two processes: a forward process trained to deconstruct the input data by adding noise and a reverse process that learns to reconstruct the data.
DALL-E 2~\cite{ramesh2022hierarchical} uses a diffusion model to learn a decoder that transforms the input data into realistic images.
For this, CLIP~\cite{radford2021learning} is used to encode images into features.
CLIP uses a contrastive objective to relate text with images, building a powerful and flexible model that can be used in applications related to natural language and visual concepts.
Taking advantage of these aspects, DALL-E 2 uses CLIP to transform an input text into an image embedding for training the diffusion model to generate images.
For this, it is trained using pairs of images and their corresponding text captions.

A model that combined CLIP with BigGAN for text-to-image generation was proposed in~\cite{bigsleep}.
A pre-trained version of both CLIP and BigGAN is used in the model.
In this case, the training process is applied for each input text, aiming to discover the best input for BigGAN that represents the selected sentence.
For this, CLIP is used to encode both the input text and generated images into embeddings.
Therefore, BigGAN receives a latent vector to create an image to be encoded through CLIP.
Then, the image embedding is compared with the encoded input text by using a distance function between the embeddings.
The training process uses backpropagation to minimize the distance and to approximate the input of BigGAN, learning to represent the selected input text according to its latent representation.
Another proposal uses a similar approach to combine VQGAN with CLIP~\cite{crowson2022vqgan}.

Evolutionary algorithms were used in~\cite{fernandes2020evolutionary} to explore the latent space of GANs.
Genetic algorithms and MAP-Elites~\cite{mouret2015illuminating} were used to navigate the latent space of GANs.

In this work, we evaluate the use of evolution strategies to search through the latent space of GANs.
Specifically, we use Covariance Matrix Adaptation Evolution Strategy (CMA-ES)~\cite{hansen2001completely}.
CMA-ES is an evolution strategy algorithm used for optimization in continuous domains.
It uses a covariance matrix to adapt the normal mutation distributions and guide the algorithm toward the objective.
The initial standard deviation $\sigma$ should be provided as input for determining the initial normal distribution.
As CMA-ES is the state-of-the-art algorithm for applying evolution strategies, and the latent space of GANs is continuous, we select it to represent the evolutionary approach in this work.

\section{Proposed Model} \label{sec:model}
We adapt the framework for text-to-image generation proposed in~\cite{bigsleep}.
The original work uses backpropagation with the Adam optimizer~\cite{kingma2015adam} in the training process.
We expand this setup by changing how the latent inputs are provided to the generator, allowing it to use not only Adam but also evolutionary algorithms such as CMA-ES.
Next, we detail the model and how the evolutionary algorithm is used.

The approach used in~\cite{bigsleep} relies on two pre-trained models: CLIP and BigGAN.
The BigGAN model is trained on the ImageNet dataset~\cite{russakovsky2015imagenet} and is used as the generative model.
CLIP is trained on $400$ million pairs of images and texts collected from the internet.

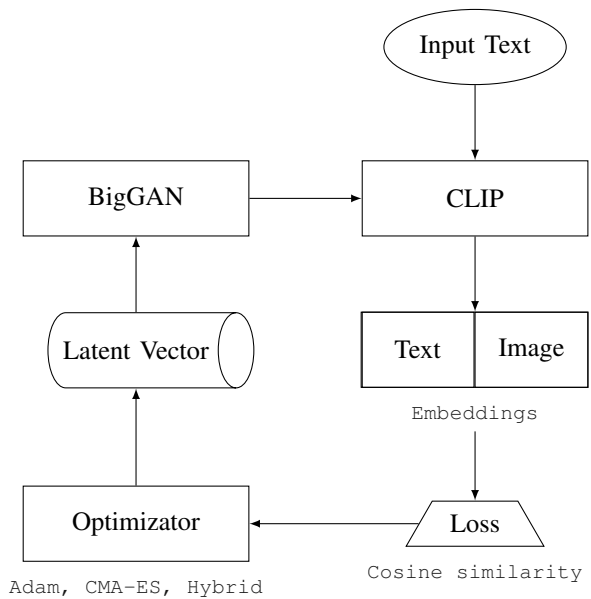
\begin{figure}[h]
	\centering
	\begin{tikzpicture}[
	node distance=8mm,
	title/.style={font=\fontsize{8}{8}\color{black!90}\ttfamily},
	block/.style= {draw, rectangle, align=center,minimum width=3cm,minimum height=1cm},
	block2/.style= {draw, rectangle, align=center,minimum width=1.5cm,minimum height=1cm},
	block3/.style= {draw, cylinder, align=center,minimum width=1.0cm,minimum height=0.7cm},
	block4/.style= {draw, trapezium, align=center,minimum width=0.6cm,minimum height=0.6cm},
	input/.style= {draw, ellipse, align=center,minimum width=2cm,minimum height=1cm},
	image/.style= {align=center,minimum width=2cm,minimum height=1cm,inner sep=0pt},
	]
	\node [block]  (clip) {CLIP};
	\node [input, above =1cm of clip] (dataset) {Input Text};
	\node [block, below =1cm of clip] (embeddings) {};
	\node [block2, left =-1.5cm of embeddings] (embeddings-text) {Text};
	\node [block2, right =0cm of embeddings-text] (embeddings-image) {Image};
	\node [title, below =0.1cm of embeddings] (embeddings-label) {Embeddings};
	\node [block, left =3cm of clip.center]  (biggan) {BigGAN};
	\node [block4, below =1.5cm of embeddings] (loss) {Loss};
	\node [title, below =0.1cm of loss] (loss-label) {Cosine similarity};
	\node [block, left =3cm of loss.center] (optimizator) {Optimizator};
	\node [title, below =0.1cm of optimizator] (optimizator-label) {Adam, CMA-ES, Hybrid};
	\node [block3, above =1.8cm of optimizator.center] (latent) {Latent Vector};

	\path[draw,->]
	(clip) edge (embeddings)
	(biggan) edge (clip)
	(dataset) edge (clip)
	(embeddings-label) edge (loss)
	(loss) edge (optimizator)
	(optimizator) edge (latent)
	(latent) edge (biggan)
	;
	\end{tikzpicture}
	\caption{Architecture of the generative model based on CLIP and BigGAN.} \label{fig:gen_model}
\end{figure}

Figure~\ref{fig:gen_model} shows an overview of the architecture of the framework.
The objective of the training process is to find the best latent inputs for BigGAN that can generate an image corresponding to the input text.
For this, CLIP is used in two moments.
Firstly, the input text is encoded to generate a feature vector with $512$ elements.
This feature vector will be used in the whole training process to compare how the generated images approximate the desired description given by the input text.
Thus, at each iteration, the input latent is updated using an optimization algorithm (e.g., Adam), and then they are fed as input to BigGAN for generating the image.
At this moment, CLIP is used again to encode the image into a feature vector.
Given these two feature vectors, i.e., one for the text and one for the image, the cosine similarity function is applied to calculate the similarity between the input text and the image.
To calculate this metric, a windowing method is used to compute the feature vector for several cuts of the generated image.
Therefore, the cosine similarity is applied with the single feature vector of the text input and every vector representing a part of the generated image.
This approach improves the comparison strategy making it more robust for detecting relevant sections of the image corresponding to the text description.

Given that the generated image plays a crucial role in the model, the inputs for BigGAN are paramount for the method presented in this work.
In BigGAN, both the class conditioning vector and the latent input can be provided not only to the input layer but also to every hidden block of the generative neural network.
We use this multiple-layer approach for the inputs, building an input vector compatible with the architecture of BigGAN, consisting of one input layer and $14$ hidden blocks.
However, as proposed in~\cite{bigsleep}, instead of providing the same input, we create one specific input for each layer.
This strategy allows for greater flexibility on the produced samples, as each layer can be influenced by its respective input vector.

When trained in ImageNet, the class conditioning vector consists of $1000$ elements, corresponding to the number of classes of the dataset.
However, to reduce the dimensionality of the input, we use the generator embedding from BigGAN to project the class conditioning vector to the same dimension of the latent input, typically set to $128$.
Thus, the input used for BigGAN consists of $(1 + |h|) \times 2 \times |z|$ elements, where $|h|$ is the number of hidden layers and $|z|$ is the latent dimension, resulting in an input of $3840$ elements.

Three approaches are used to optimize this input for the desired text-to-image task: backpropagation with Adam, CMA-ES, and a hybrid solution.

The backpropagation approach to optimize the input array uses the Adam optimizer to adjust the vector for the target.
In this case, the evaluation function given by the cosine similarity between the feature spaces of the text and the image is used as the loss function for backpropagation and to approximate the input vector to the desired input text through gradient descent.

For the evolutionary strategy, we use the input of $3840$ elements as the individuals for CMA-ES.
Therefore, each individual represents a latent vector that will produce a specific image when provided to BigGAN.
In this way, the CMA-ES algorithm produces new generations of individuals according to its mutation strategy, being guided by the fitness function that uses the cosine similarity calculation method.

In the hybrid approach, we apply $k$ steps of the backpropagation method before applying the mutation step from CMA-ES.
In this way, we make use of the two strategies to approximate the latent vector to the desired target.

\section{Experiments}
\subsection{Evaluation Method} \label{sec:eval}
We adapt the evaluation method proposed in~\cite{costa2021tsne} to analyze the distribution of samples produced in our experiments~\footnote{Code available at \url{https://github.com/vfcosta/gen-tsne}.}.
Originally, the evaluation method used the components of a GAN in two steps of the evaluation process.
The generator is used to produce synthetic samples and the discriminator is used to extract features of these samples and also from the input dataset, providing the inputs for the t-SNE algorithm.
The input dataset is used as a baseline to compute the Jaccard Index to quantify how much the generative models capture the distribution.

The number of iterations and perplexity are two configuration parameters for the t-SNE algorithm.
Perplexity defines how the neighborhood of each data point is handled and is usually defined by values in the range of $5$ to $50$~\cite{maaten2008visualizing}.
The number of iterations limits the number of steps used to calculate the projections in the algorithm.

In this work, we propose to adapt the architecture of the evaluation method to use our generative models to synthesize the samples and CLIP as the feature extractor.
As we do not rely on an input dataset, we use the best-performing method as the baseline to compute the relative performance of the other methods.

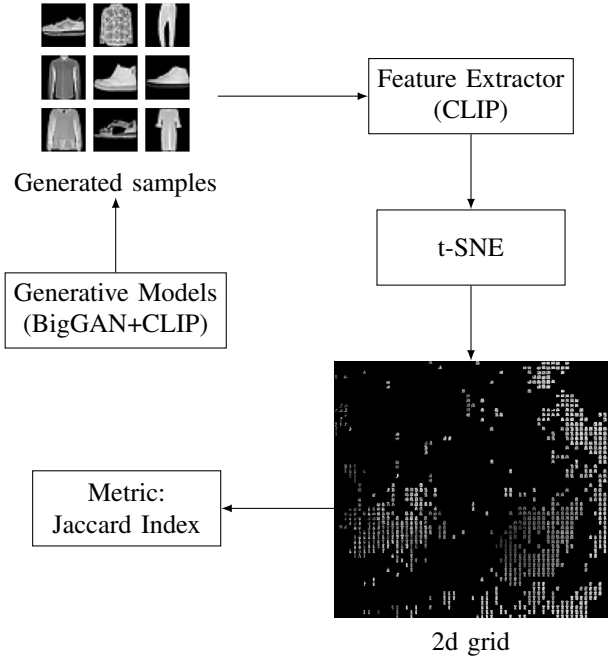
\begin{figure}[h]
	\centering
	\begin{tikzpicture}[
	node distance=8mm,
	title/.style={font=\fontsize{8}{8}\color{black!80}\ttfamily},
	block/.style= {draw, rectangle, align=center,minimum width=2.5cm,minimum height=1cm},
	image/.style= {align=center,minimum width=2cm,minimum height=1cm,inner sep=0pt},
	typetag/.style={rectangle, draw=black!50, font=\scriptsize\ttfamily, anchor=west}
	]
	\node [block]  (discriminator) {Feature Extractor\\(CLIP)};
	\node [image, left =2cm of discriminator] (dataset) {\includegraphics[width=.12\textwidth]{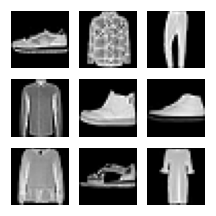}\\Generated samples};
	\node [block, below =1cm of dataset]  (generator) {Generative Models\\(BigGAN+CLIP)};
	\node [block, below =1cm of discriminator] (tsne) {t-SNE};
	\node [image, below =1cm of tsne] (tsneimage) {\includegraphics[width=.2\textwidth]{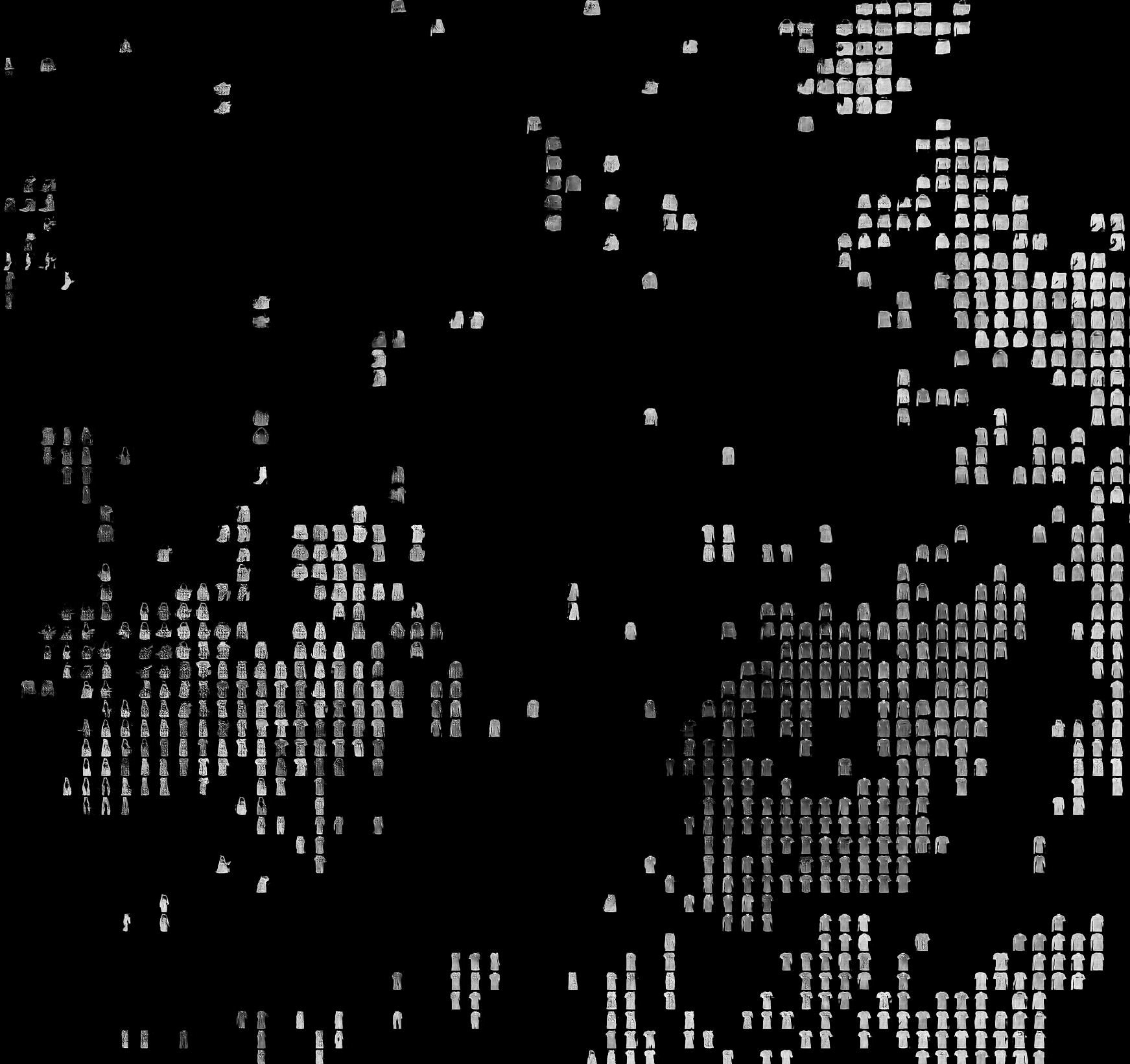}\\2d grid};
	\node [block, left =1.5cm of tsneimage] (metrics) {Metric:\\Jaccard Index};
	
	\path[draw,->]
	(discriminator) edge (tsne)
	(generator) edge (dataset)
	(dataset) edge (discriminator)
	(tsne) edge (tsneimage)
	(tsneimage) edge (metrics)
	;
	\end{tikzpicture}
	\caption{Overview of the evaluation method proposed in this work to analyze the distribution of the images produced by the generative models.} \label{fig:eval_method}
\end{figure}

\subsection{Experimental Setup}
We evaluate the approach described in Section~\ref{sec:model} using three different approaches to explore the latent space: Backpropagation with the Adam optimizer, an Evolutionary Algorithm through CMA-ES, and the Hybrid approach that combines Adam and CMA-ES.
The experiments are conducted considering the generation of images to the following sentences:
\begin{enumerate}
	\item A painting of Superman by Van Gogh;
	\item A painting of a fox in the style of Starry Night;
	\item A painting of Darth Vader in the style of Matisse.
\end{enumerate}

Our choice for the text inputs is focused on art generation, also using popular elements of contemporary culture.

\begin{table}[ht]
	\caption{Experimental Parameters}
	\label{table:setup}
	\centering
	\begin{tabular}{c c}
		\hline\hline
		\textbf{Evolutionary Parameters} & \textbf{Value} \\
		\hline
		Number of Generations (CMA-ES) & $100$ \\
		Number of Generations (Hybrid) & $50$ \\
		Population size & $10$ \\
		$\sigma$ & $0.2$ \\
		\textbf{Backpropagation Parameters} & \textbf{Value} \\
		\hline
		Iterations & $1000$ \\
		Optimizer & Adam \\
		Learning rate & 0.05 \\
		Latent input & 128 \\
		\textbf{Evaluation Parameters} & \textbf{Value} \\
		\hline
		t-SNE Perplexity & 40 \\
		t-SNE Iterations & 1000 \\
		Samples per model & 500 \\
		Executions & 30 \\
		\hline\hline
	\end{tabular}
\end{table}

Table \ref{table:setup} lists the parameters used in our experiments.
The CMA-ES approach uses a population of $10$ individuals and $\sigma = 0.2$.
The gradient-based approach uses Adam as the optimizer with a learning rate of $0.05$.
For the evaluation method, we use perplexity of $40$ and $1000$ iterations for the t-SNE algorithm.
The calculation of the Jaccard Index is repeated for $30$ executions to take into account the non-determinism behavior of the t-SNE algorithm.

Each scenario was repeated for $500$ executions.
Images from the last iteration of each execution are used with the evaluation method described in this section for analyzing the results.

To make the comparison fair, we run the same number of evaluations for each approach.
We run the Adam version for $1000$ iterations.
As CMA-ES uses a population of $10$ individuals, we run this approach for $100$ generations.
Thus, the CMA-ES will execute $1000$ evaluations by using the cosine similarity for guiding the evolution of the individuals.
The hybrid approach also uses $10$ individuals, but we run this approach for $50$ generations to take into account the double number of evaluations given by the application of one step of Adam ($k = 1$) and the CMA-ES evaluations.

\section{Results}
\begin{figure*}[htb]
	\centering
	\subfloat[Adam]{\includegraphics[width=0.32\linewidth]{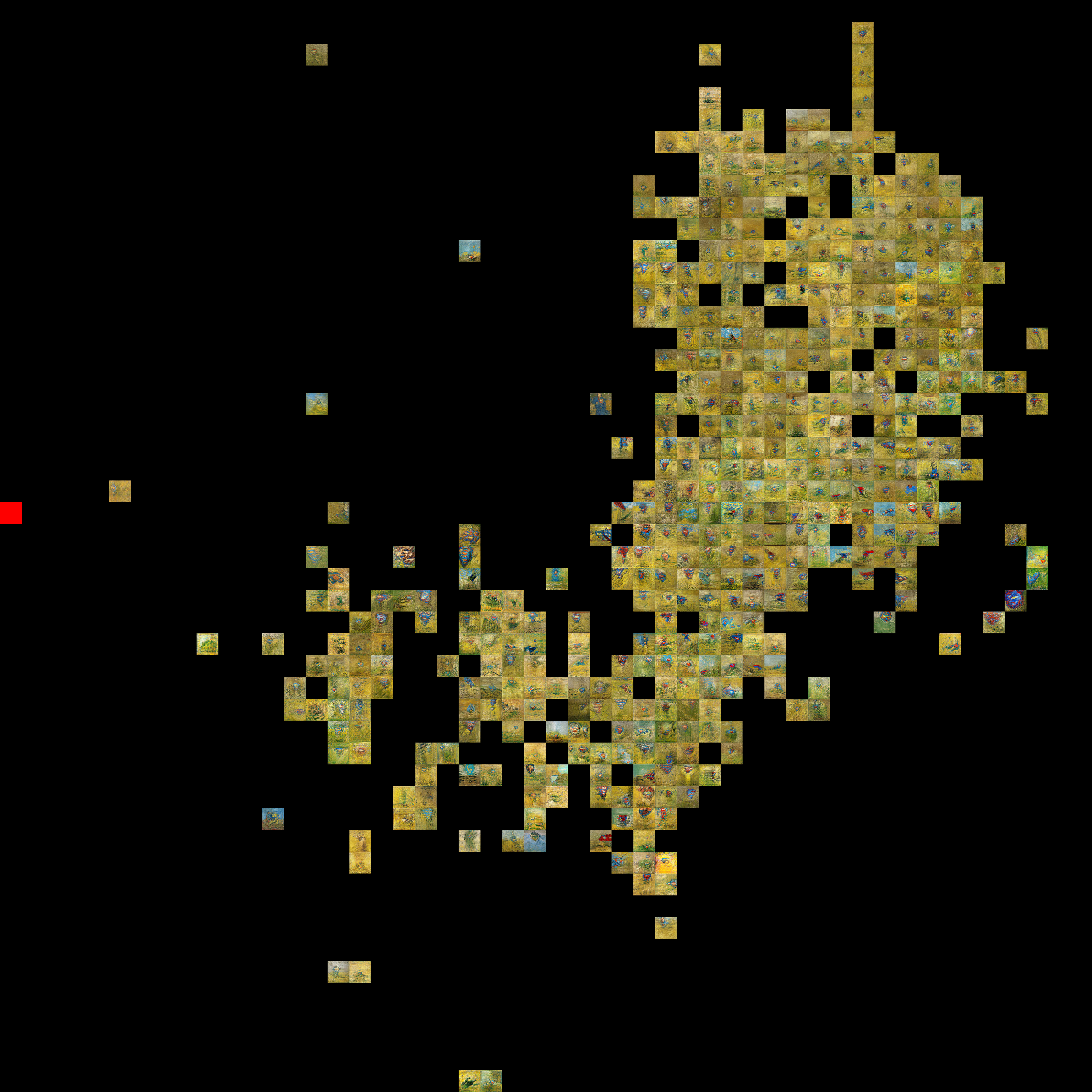}%
		\label{fig:tsne_super_adam}}
	\hfil
	\subfloat[CMA-ES]{\includegraphics[width=0.32\linewidth]{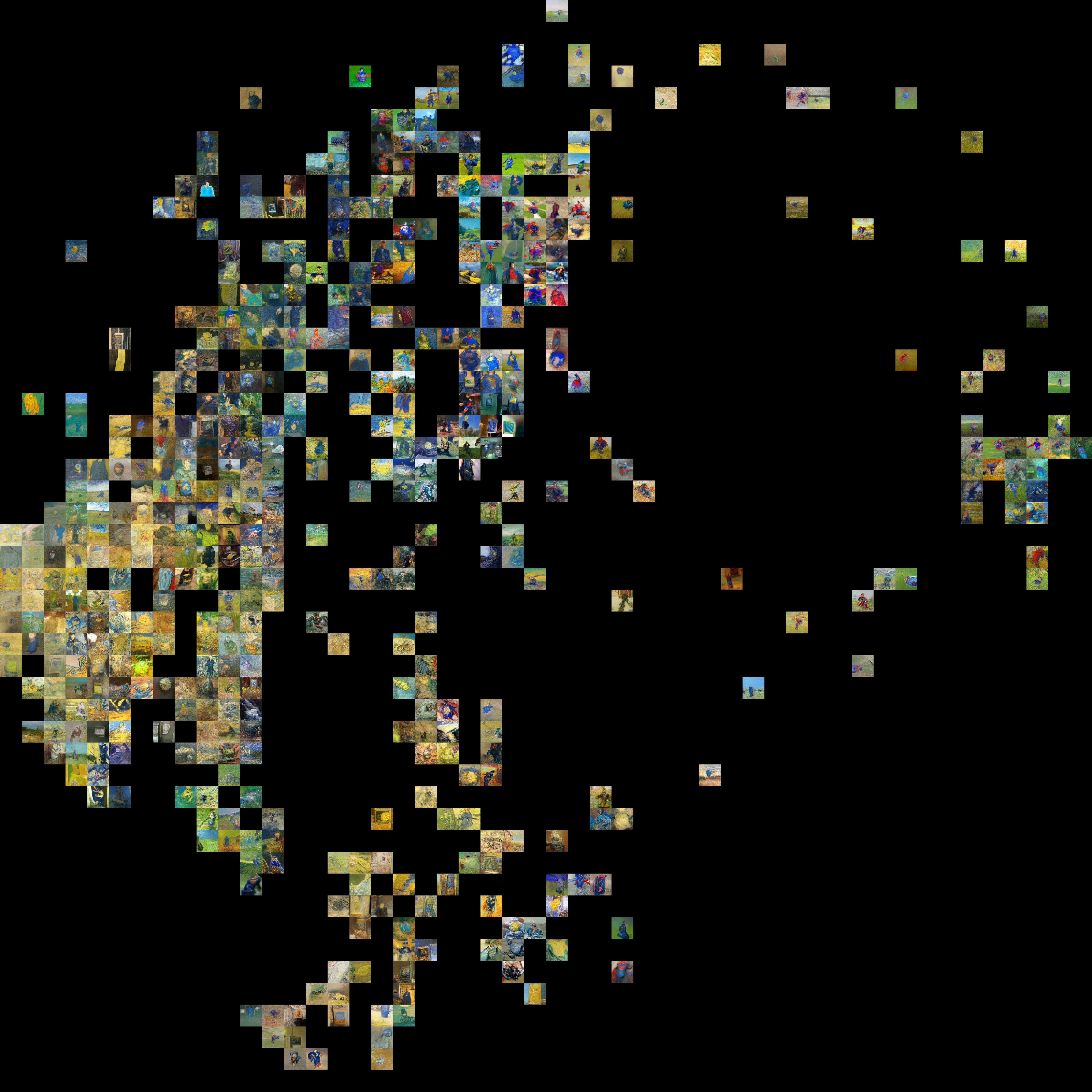}%
		\label{fig:tsne_super_cmaes}}
	\hfil
	\subfloat[Hybrid]{\includegraphics[width=0.32\linewidth]{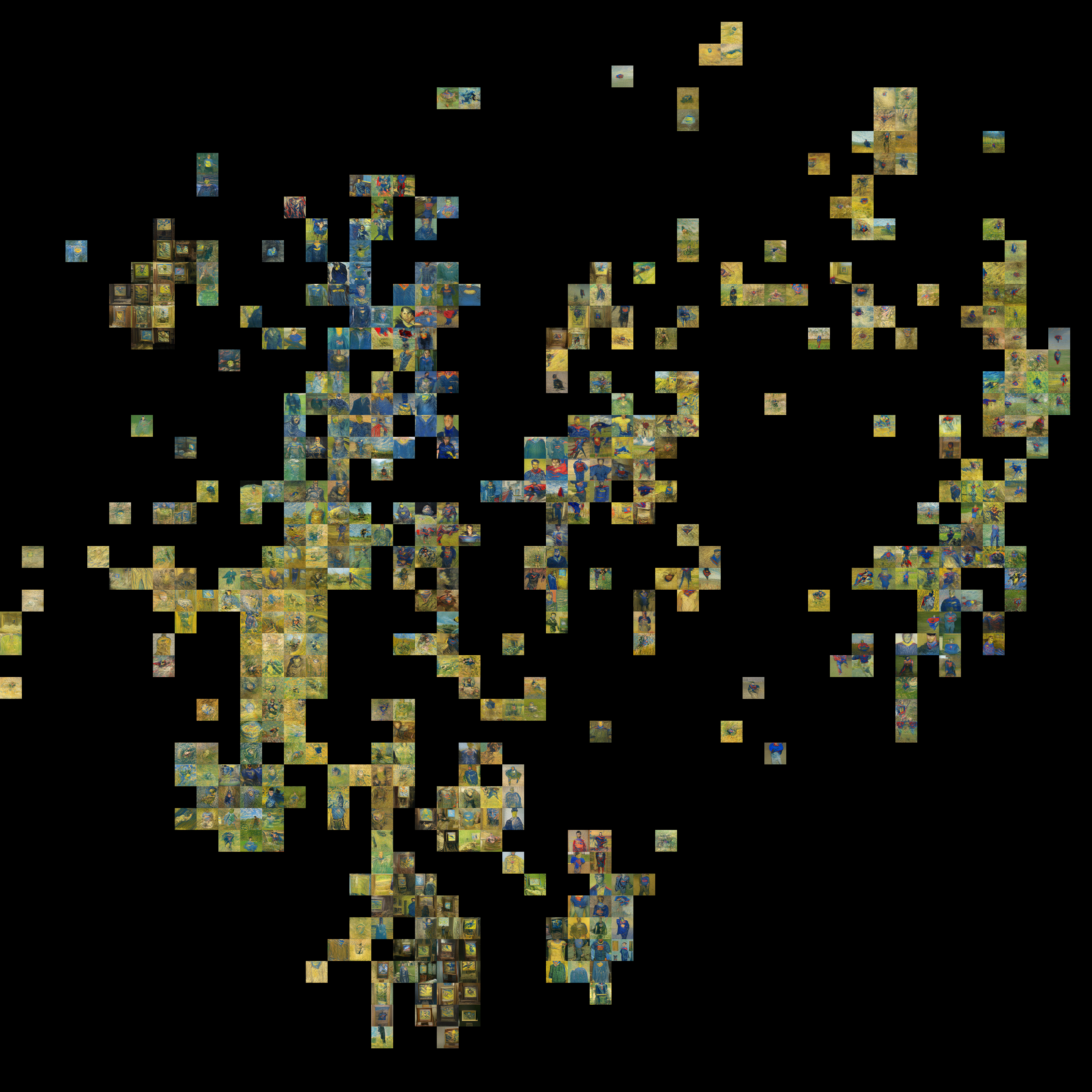}%
		\label{fig:tsne_super_hybrid}}
	\hfil
	\caption{Two-dimensional grid revealing the distribution of images generated using Adam~\subref{fig:tsne_super_adam}, CMA-ES~\subref{fig:tsne_super_cmaes}, and the Hybrid approach~\subref{fig:tsne_super_hybrid} for the text input "A painting of Superman by Van Gogh" (images with full resolution available at https://github.com/vfcosta/clip-gan-es/tree/main/images/superman).}
	\label{fig:tsne_super}
\end{figure*}

%
Figure~\ref{fig:tsne_super} presents the distribution of images in the two-dimensional grid produced by the evaluation method using the first text input: "A painting of Superman by Van Gogh".

\begin{figure*}[htb]
	\centering
	\subfloat[Adam]{\includegraphics[width=0.32\linewidth]{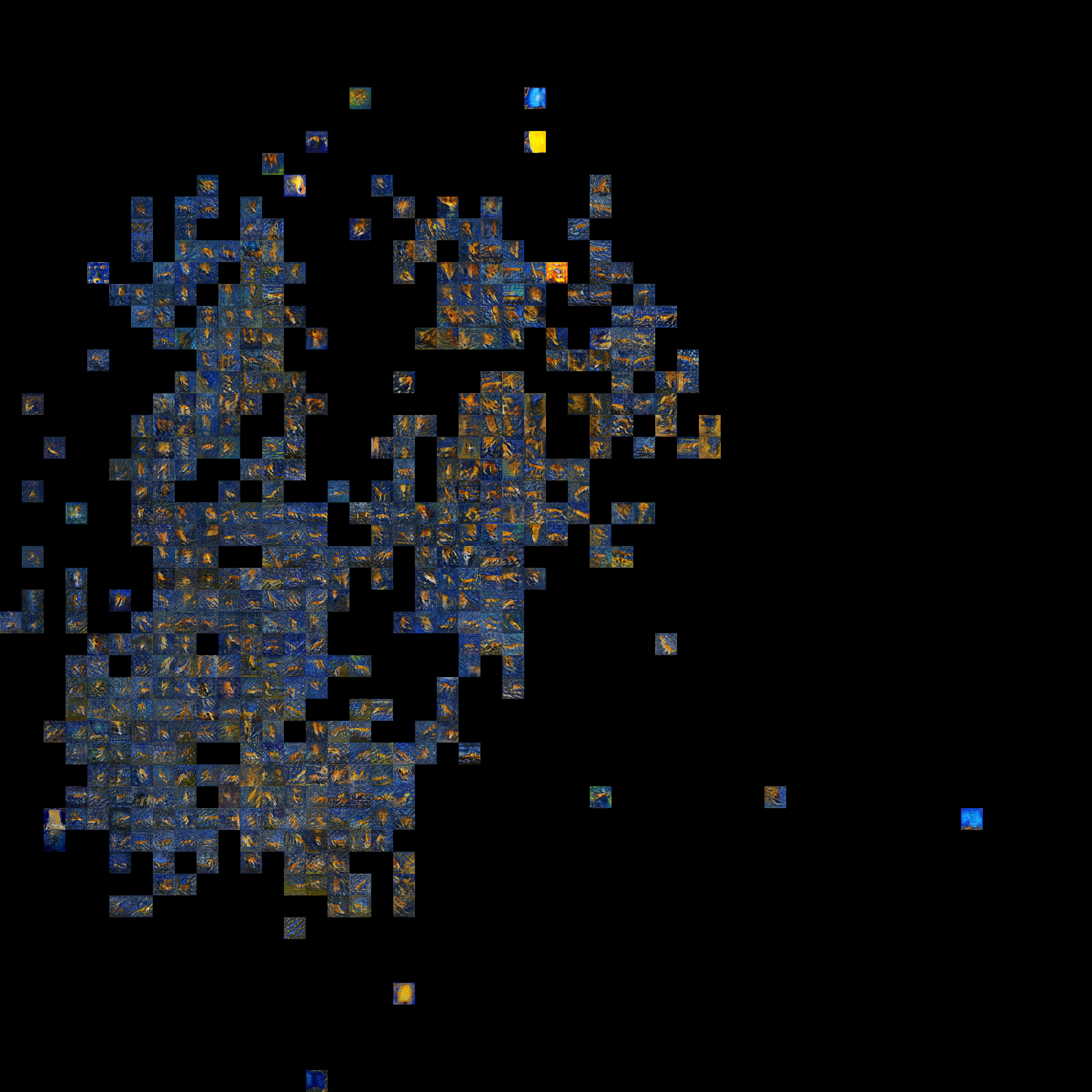}%
		\label{fig:tsne_fox_adam}}
	\hfil
	\subfloat[CMA-ES]{\includegraphics[width=0.32\linewidth]{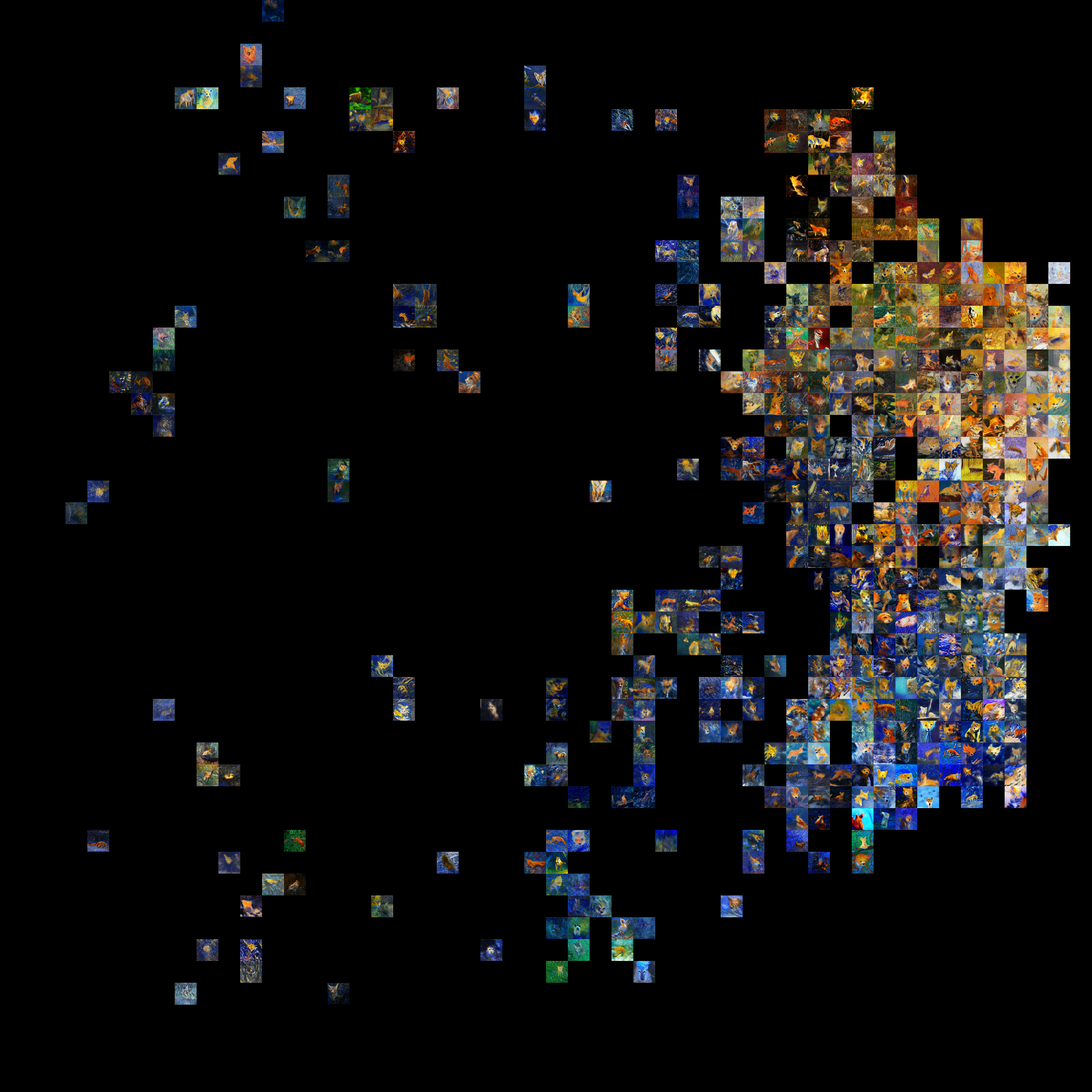}%
		\label{fig:tsne_fox_cmaes}}
	\hfil
	\subfloat[Hybrid]{\includegraphics[width=0.32\linewidth]{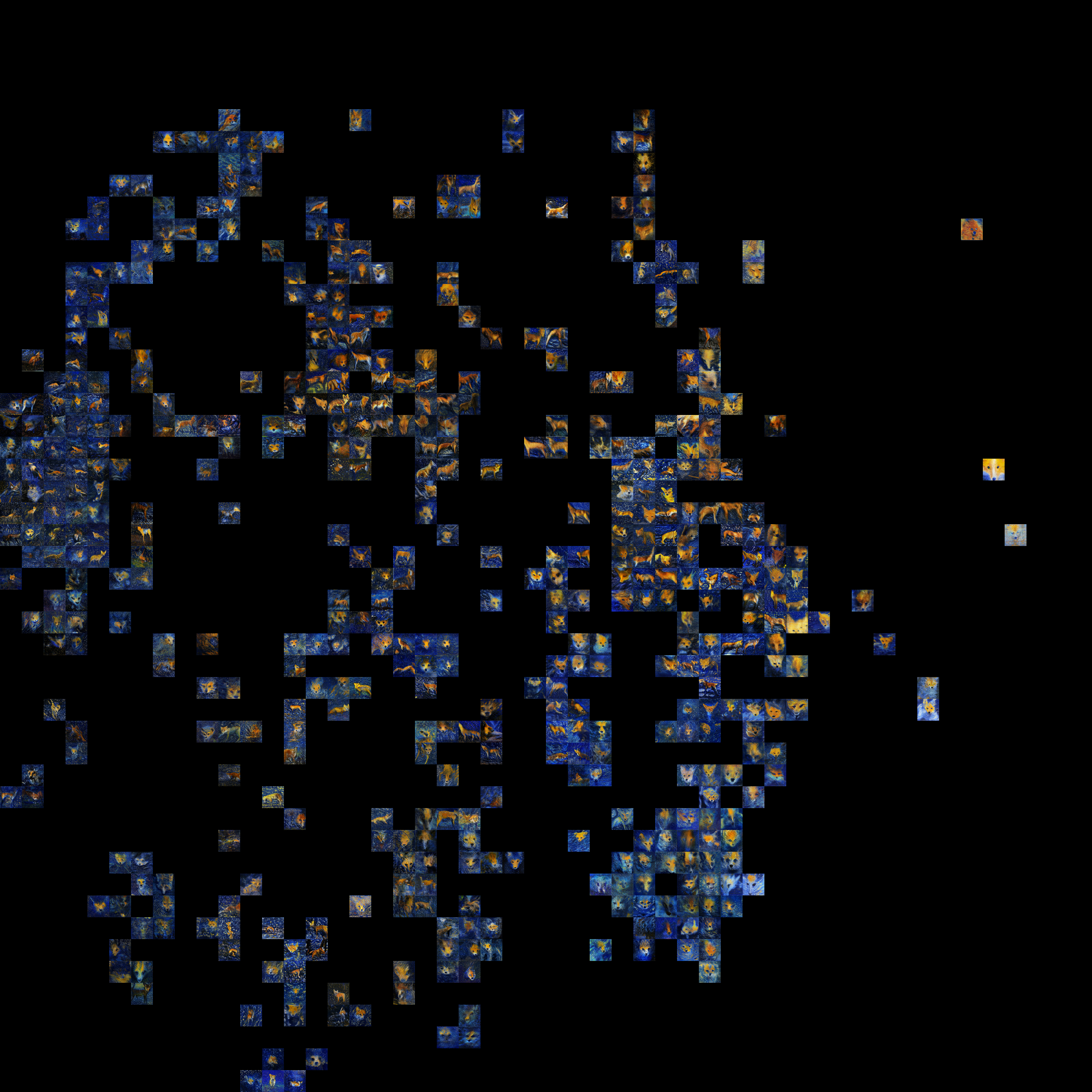}%
		\label{fig:tsne_fox_hybrid}}
	\hfil
	\caption{Two-dimensional grid revealing the distribution of images generated using Adam~\subref{fig:tsne_fox_adam}, CMA-ES~\subref{fig:tsne_fox_cmaes}, and the Hybrid approach~\subref{fig:tsne_fox_hybrid} for the text input "A painting of a fox in the style of Starry Night" (images with full resolution available at https://github.com/vfcosta/clip-gan-es/tree/main/images/fox).}
	\label{fig:tsne_fox}
\end{figure*}

The first grid (Figure~\ref{fig:tsne_super_adam}) contains the images generated by BigGAN when using Adam to optimize the latent space.
Looking at the grid, we can see that images are concentrated in a specific region, indicating that the diversity of the created samples is limited when using this approach.
When inspecting the next strategy, namely CMA-ES displayed in Figure~\ref{fig:tsne_fox_cmaes}, we can see that the images are more distributed through the grid.
This is an indication that the CMA-ES is promoting a better exploration of the search space, allowing for a more diverse set of results.
Another interesting observation emerges when we compare the CMA-ES results with Adam’s.
In concrete, we can see that they are exploring non-overlapping regions, indicating that they follow different paths when performing the optimization of the latent space.
When combining Adam with CMA-ES, we leverage the results from both approaches.
Looking at the results from Figure~\ref{fig:tsne_fox_hybrid}, we can see a mixture of the results of both approaches.

To confirm our observations, we apply the Jaccard Index using the projected grids as proposed in ~\cite{costa2021tsne}.
The Jaccard Index measures the intersection over union of the samples from the distributions.
For this, we use the best performing method as the baseline, i.e., the Hybrid approach.
Given this baseline, the results for the Jaccard Index for the Adam and CMA-ES approaches are $0.2978\pm0.0111$ and $0.3710\pm0.0124$, respectively.
As bigger values indicate that the distribution is closer to the baseline, our results evidence that the CMA-ES approach better approximates the Hybrid method.
The Adam version of the method achieved lower performance, evidencing the poor capacity of achieving diversity in the exploration of the latent space.
However, the Hybrid method is able to combine the two distributions, leading to a broader exploration of the search space.

\begin{figure}[htb]
	\centering
	\subfloat{\includegraphics[width=0.25\linewidth]{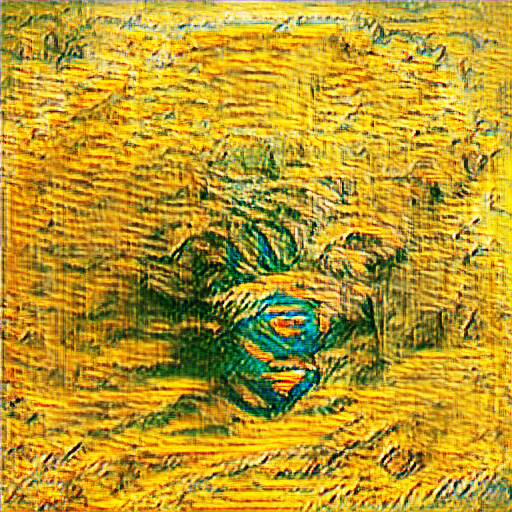}}
	\hfil
	\subfloat{\includegraphics[width=0.25\linewidth]{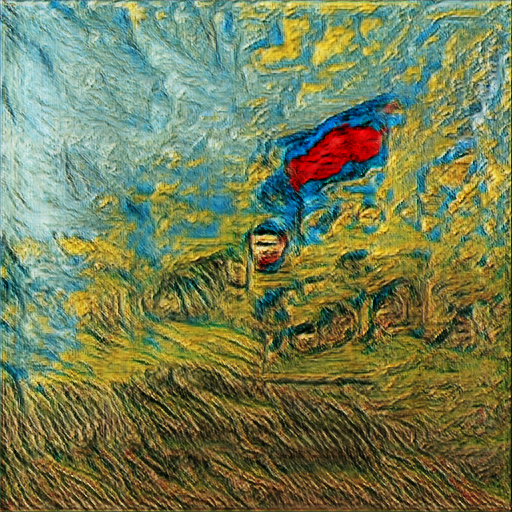}}
	\hfil
	\subfloat{\includegraphics[width=0.25\linewidth]{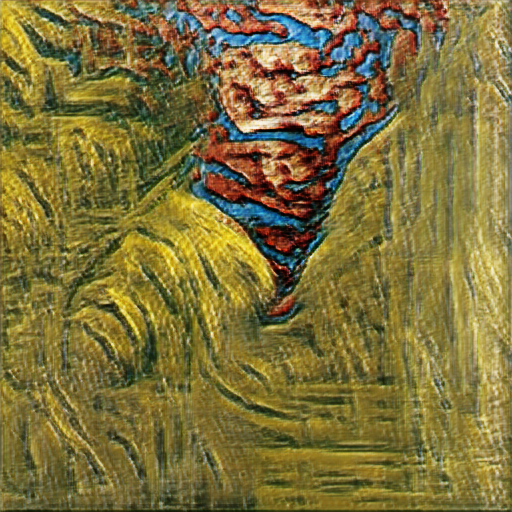}}
	\hfil
	\subfloat{\includegraphics[width=0.25\linewidth]{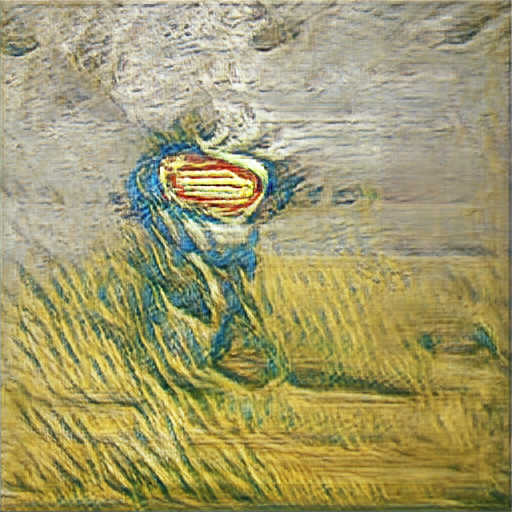}%
	\label{fig:samples_super_adam}}
	\hfil
	\subfloat{\includegraphics[width=0.25\linewidth]{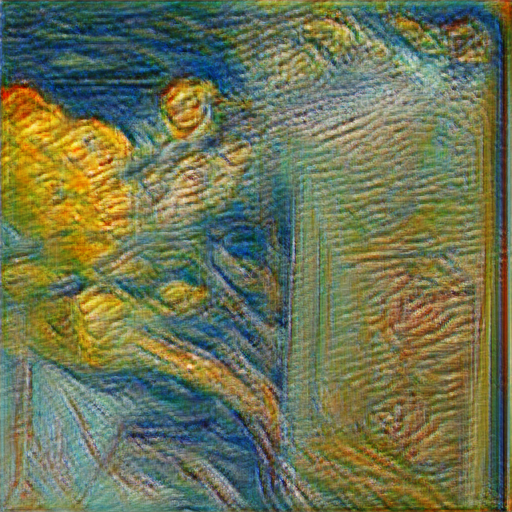}}
	\hfil
	\subfloat{\includegraphics[width=0.25\linewidth]{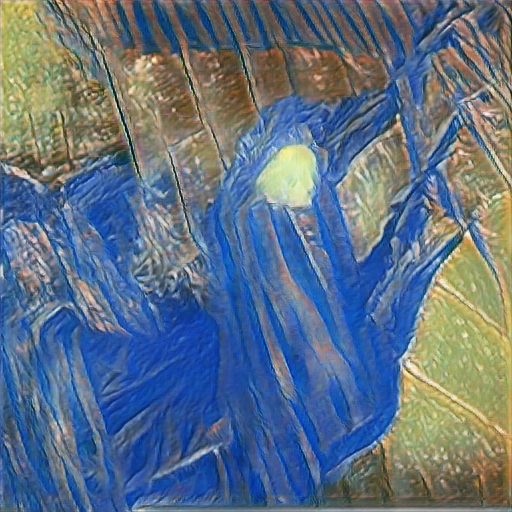}}
	\hfil
	\subfloat{\includegraphics[width=0.25\linewidth]{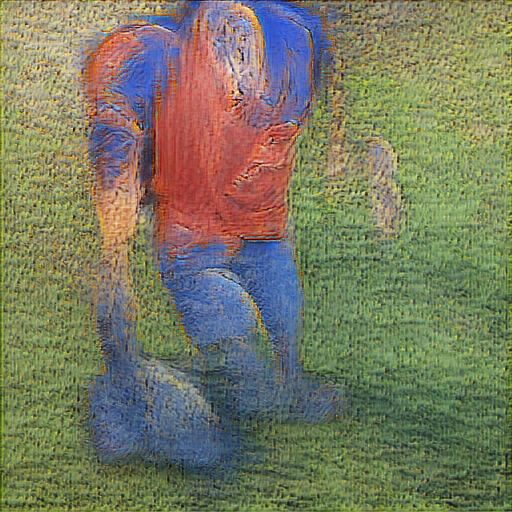}}
	\hfil
	\subfloat{\includegraphics[width=0.25\linewidth]{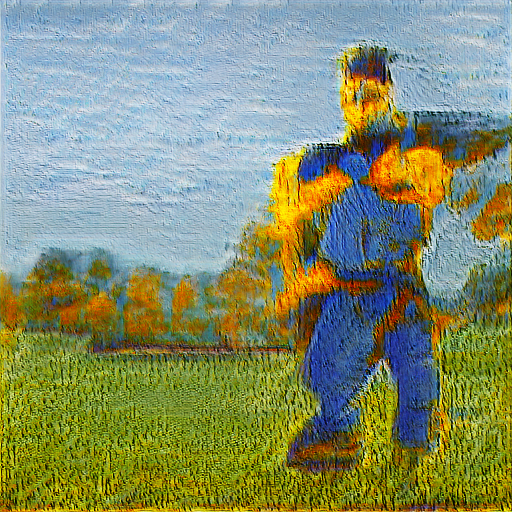}%
	\label{fig:samples_super_evo}}
	\hfil
	\subfloat{\includegraphics[width=0.25\linewidth]{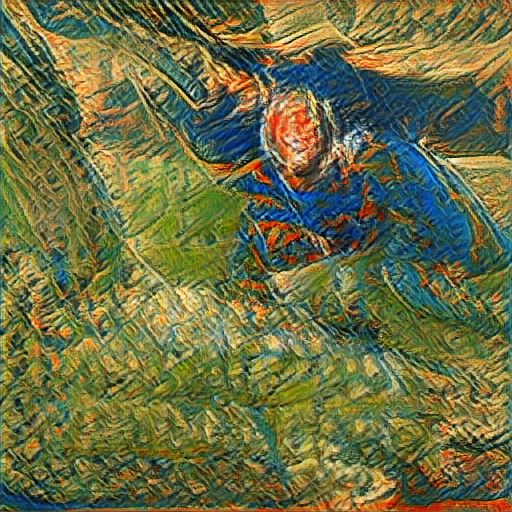}}
	\hfil
	\subfloat{\includegraphics[width=0.25\linewidth]{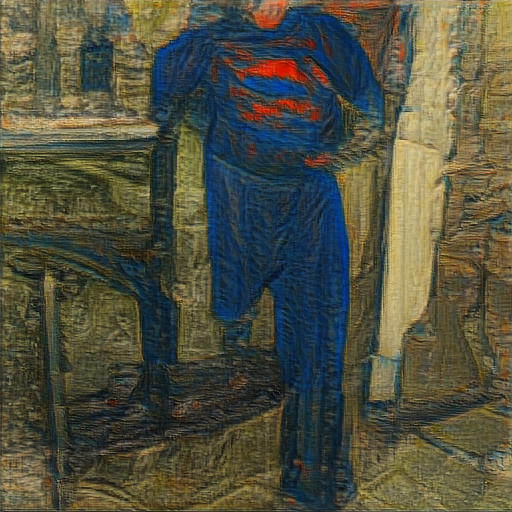}}
	\hfil
	\subfloat{\includegraphics[width=0.25\linewidth]{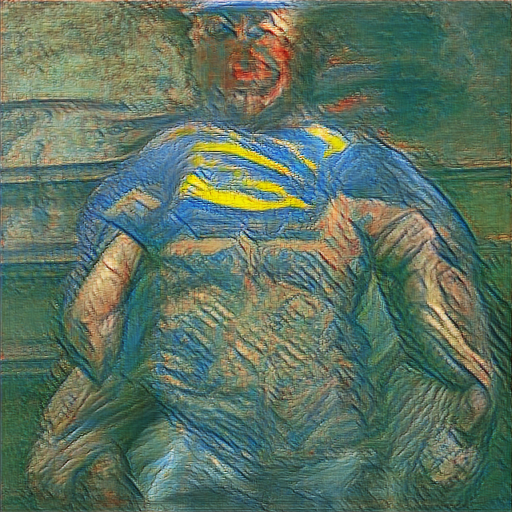}}
	\hfil
	\subfloat{\includegraphics[width=0.25\linewidth]{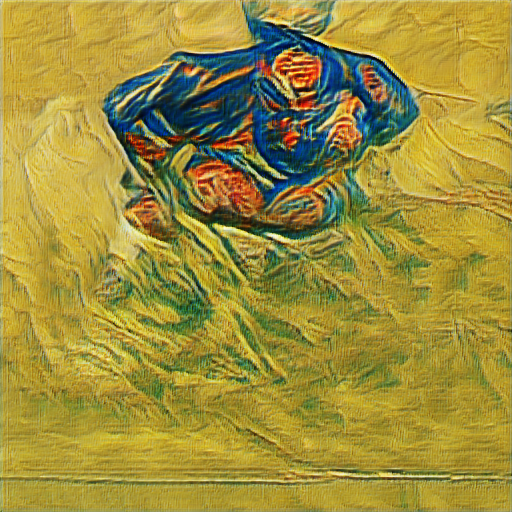}%
	\label{fig:samples_super_hybrid}}
	\hfil

	\caption{Samples from Adam (top row), CMA-ES (middle row), and the Hybrid (bottom row) approach for the text input "A painting of Superman by Van Gogh".}
	\label{fig:samples_super}
\end{figure}

In Figure~\ref{fig:samples_super}, we present some samples created using the three different methods.
Although the samples are not perfect, we can see elements from the input text in all images, such as texture and colors.
It is important to note that we use BigGAN trained on the ImageNet dataset in our experiments.
Thus, the generative model is not trained specifically on the tasks given through the input texts.
However, even in this scenario, we can see that the model is capable of representing some characteristics.

\begin{figure}[bth]
	\includegraphics[width=1\linewidth]{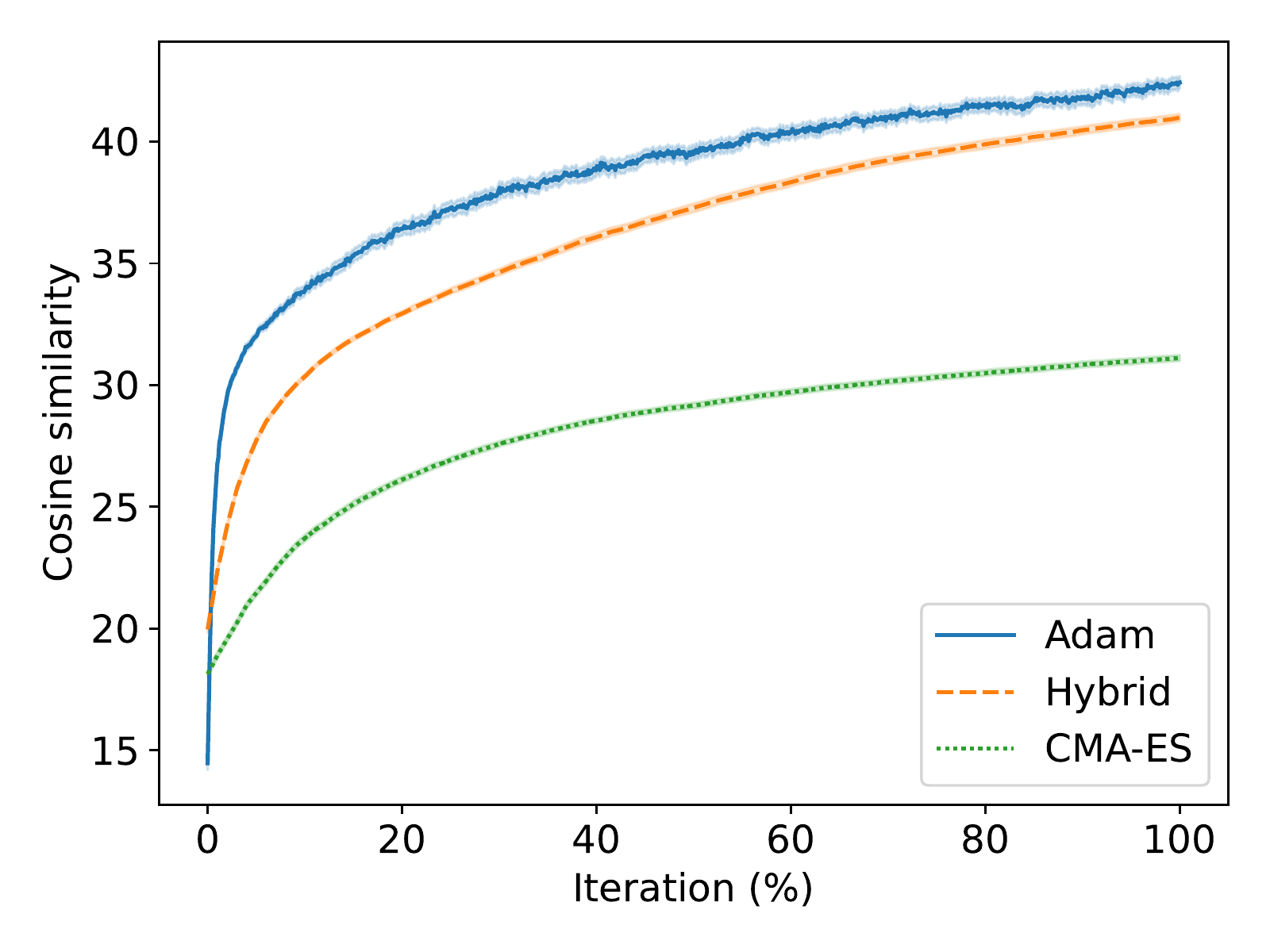}
	\caption{The cosine similarity for best images generated using Adam, CMA-ES, and the Hybrid approach for the text input "A painting of Superman by Van Gogh".}\label{fig:fitness_super}
\end{figure}

Figure~\ref{fig:fitness_super} shows the cosine similarity for the best images generated during the latent space exploration using Adam, CMA-ES, and the Hybrid approach for the text input "A painting of Superman by Van Gogh".
As we have a different number of iterations and generations for each approach, we represent on the horizontal axis the iteration percentage.
The values reported in this chart are the average cosine similarities for each approach during $500$ executions with a confidence interval of $95\%$.
The results show that the Adam approach achieved the best results regarding this metric.
However, this behavior does not correspond to the visual inspection of the images.
The CMA-ES solution achieved the lowest results, but the samples appear to be better than the Adam version.
The Hybrid approach is closer to the Adam results, making it to get the benefits of both approaches regarding the quality and diversity of the generated samples.
This shows evidence that relying on the similarity metric alone might hinder the quality of the results in terms of diversity.

As can be seen in Figure~\ref{fig:tsne_fox}, the behavior displayed in Figure~\ref{fig:tsne_super} is even more evident in the experiments with the second text input: "A painting of a fox in the style of Starry Night".

The distributions show a more evident separation between Adam, CMA-ES, and the hybrid approach.
Namely, we can see that the CMA-ES approach explores an area of the grid that is never explored by Adam.
Looking at the Hybrid approach, it is possible to see that it achieved an equilibrium between the Adam and CMA-ES search spaces.
Considering the Hybrid approach as the baseline, the results for the Jaccard Index for the Adam and CMA-ES approaches are $0.5004\pm0.0146$ and $0.1999\pm0.0093$, respectively.

\begin{figure}[htb]
	\centering
	\subfloat{\includegraphics[width=0.25\linewidth]{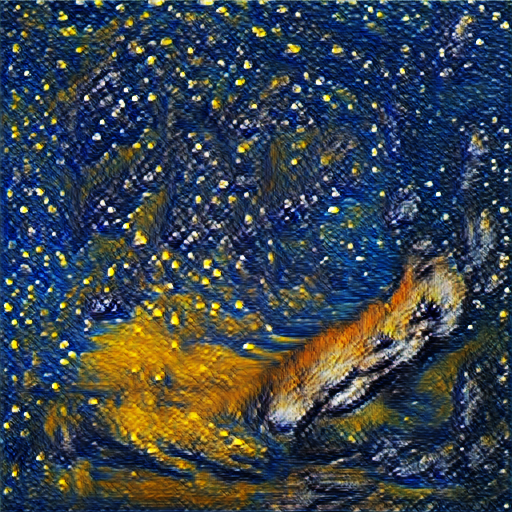}}
	\hfil
	\subfloat{\includegraphics[width=0.25\linewidth]{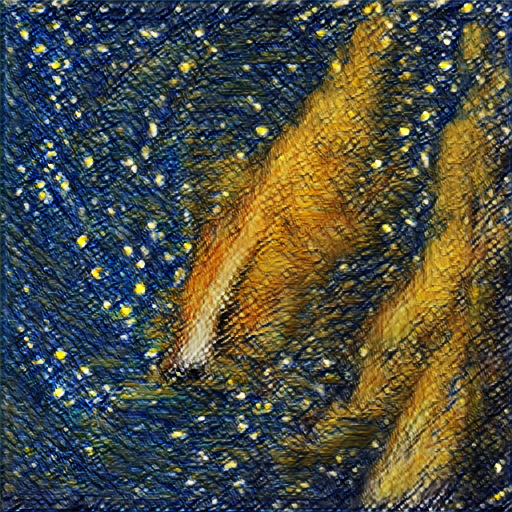}}
	\hfil
	\subfloat{\includegraphics[width=0.25\linewidth]{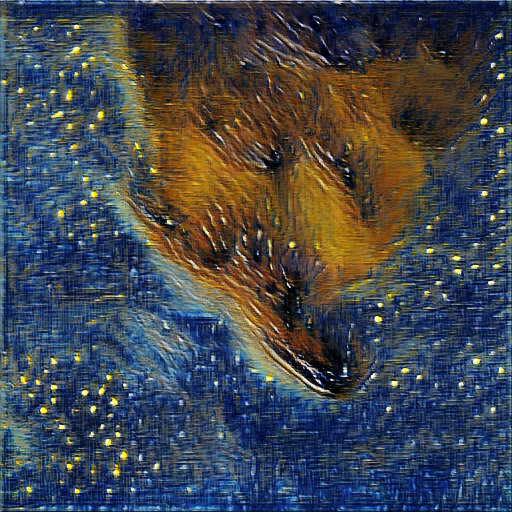}}
	\hfil
	\subfloat{\includegraphics[width=0.25\linewidth]{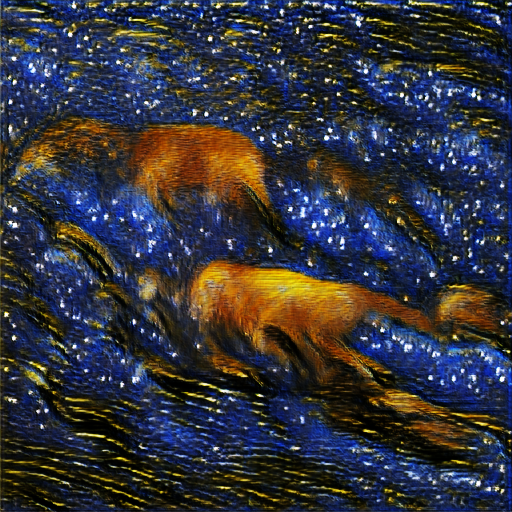}%
		\label{fig:samples_fox_adam}}
	\hfil
	\subfloat{\includegraphics[width=0.25\linewidth]{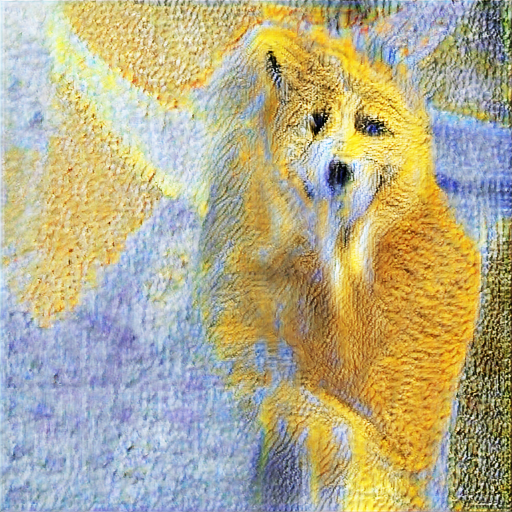}}
	\hfil
	\subfloat{\includegraphics[width=0.25\linewidth]{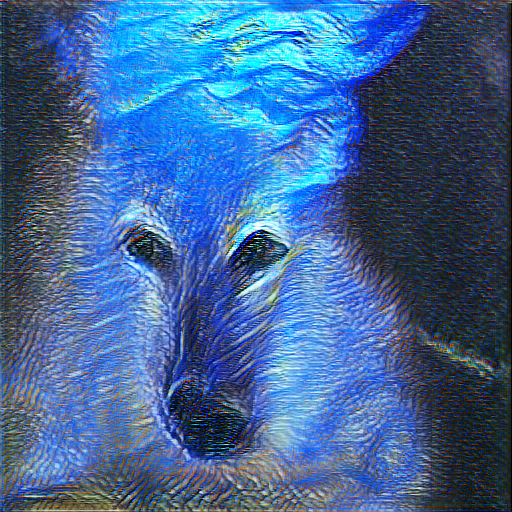}}
	\hfil
	\subfloat{\includegraphics[width=0.25\linewidth]{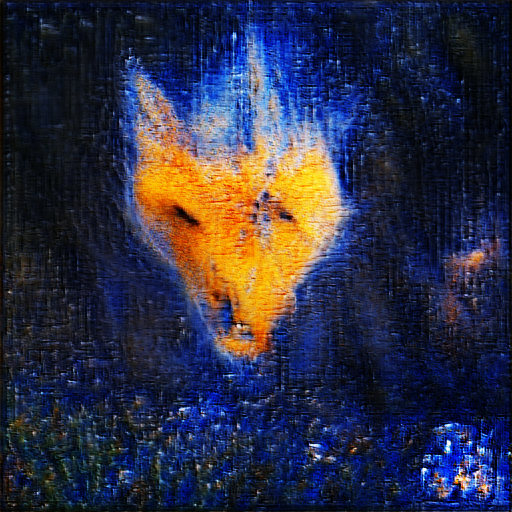}}
	\hfil
	\subfloat{\includegraphics[width=0.25\linewidth]{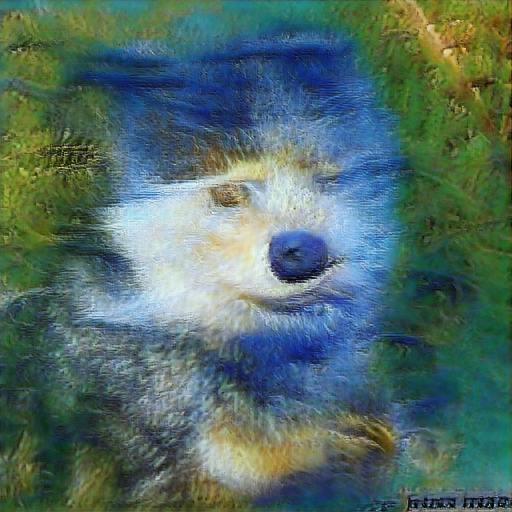}%
		\label{fig:samples_fox_evo}}
	\hfil
	\subfloat{\includegraphics[width=0.25\linewidth]{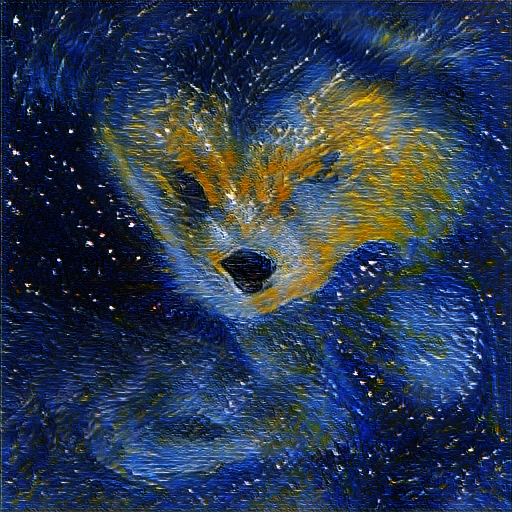}}
	\hfil
	\subfloat{\includegraphics[width=0.25\linewidth]{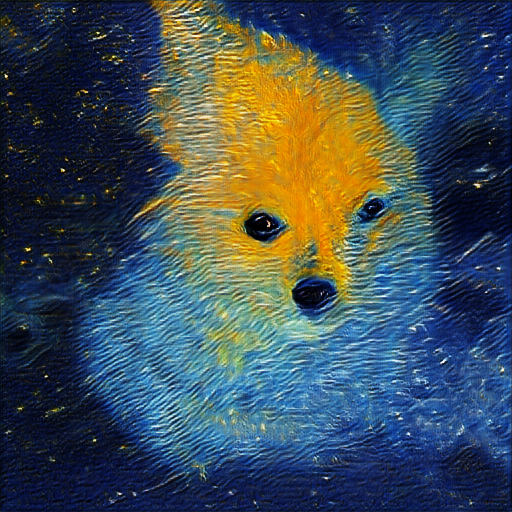}}
	\hfil
	\subfloat{\includegraphics[width=0.25\linewidth]{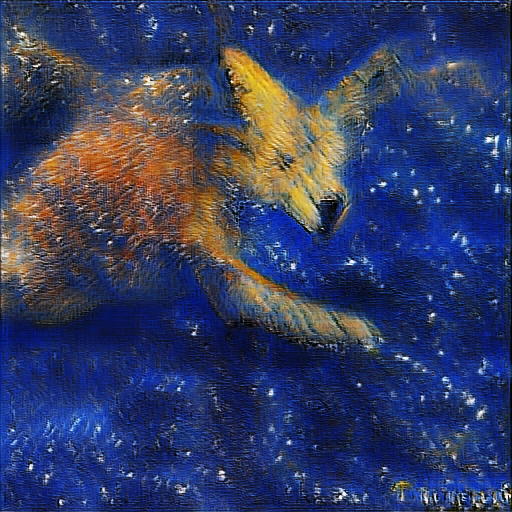}}
	\hfil
	\subfloat{\includegraphics[width=0.25\linewidth]{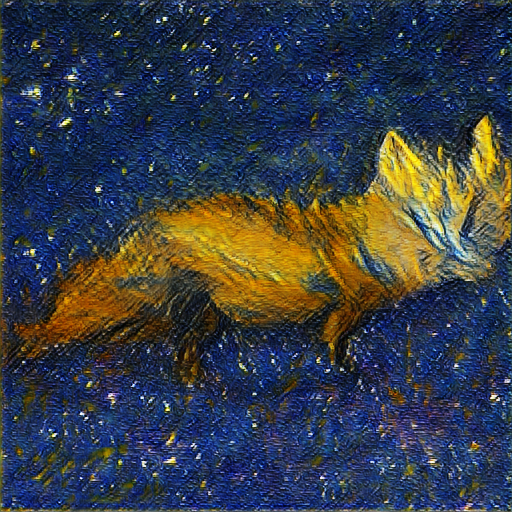}%
		\label{fig:samples_fox_hybrid}}
	\hfil
	
	\caption{Samples from Adam (top row), CMA-ES (middle row), and the Hybrid (bottom row) approach for the text input "A painting of a fox in the style of Starry Night".}
	\label{fig:samples_fox}
\end{figure}

We highlight in Figure~\ref{fig:samples_fox} samples achieved from each strategy applied in our experiments.
The top, middle, and bottom rows contain samples generated by applying Adam, CMA-ES, and the Hybrid approach.
As indicated in the text, the samples present elements resembling a fox, such as color shapes and, in some cases, a more reliable representation.
Besides, elements of the painting "The Starry Night" from Vincent van Gogh are also present, such as the blue sky and shapes resembling stars.

\begin{figure}[bth]
	\includegraphics[width=1\linewidth]{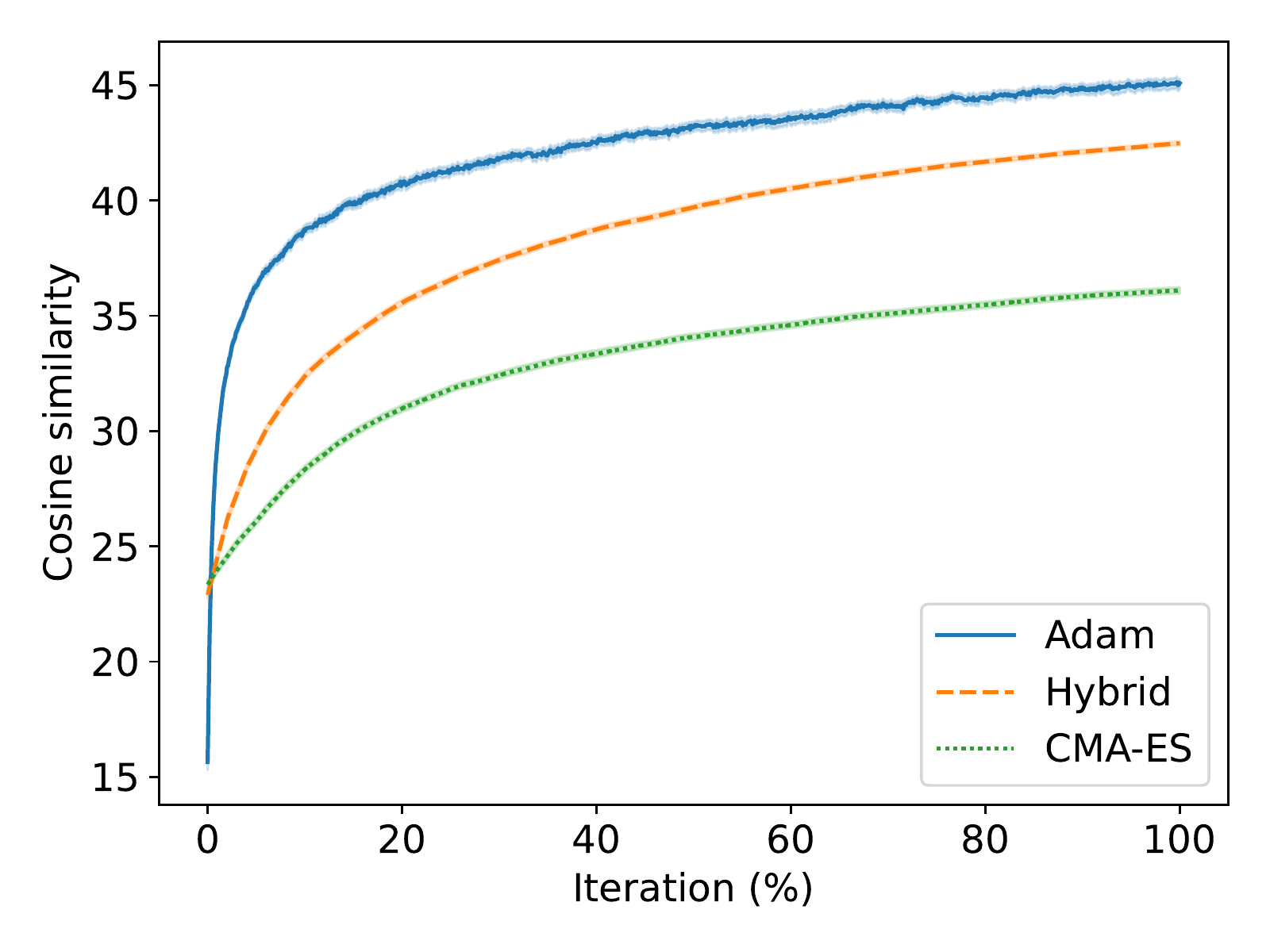}
	\caption{The cosine similarity for best images generated using Adam, CMA-ES, and the Hybrid approach for the text input "A painting of a fox in the style of Starry Night".}\label{fig:fitness_fox}
\end{figure}

Figure~\ref{fig:fitness_fox} shows the cosine similarity for the best images generated during the latent space exploration using Adam, CMA-ES, and the Hybrid approach for the text input "A painting of a fox in the style of Starry Night".
The results are similar to the behavior displayed in Figure~\ref{fig:fitness_super}.
The Adam approach produces a better cosine similarity.
However, from the visual inspection of the results, it is clear that the quality of the samples does not correspond to this evidence.
Once again, this is an indication that the metric used to compare the features of the image and the input text can be improved to better represent the similarity.

\begin{figure*}[htb]
	\centering
	\subfloat[Adam]{\includegraphics[width=0.32\linewidth]{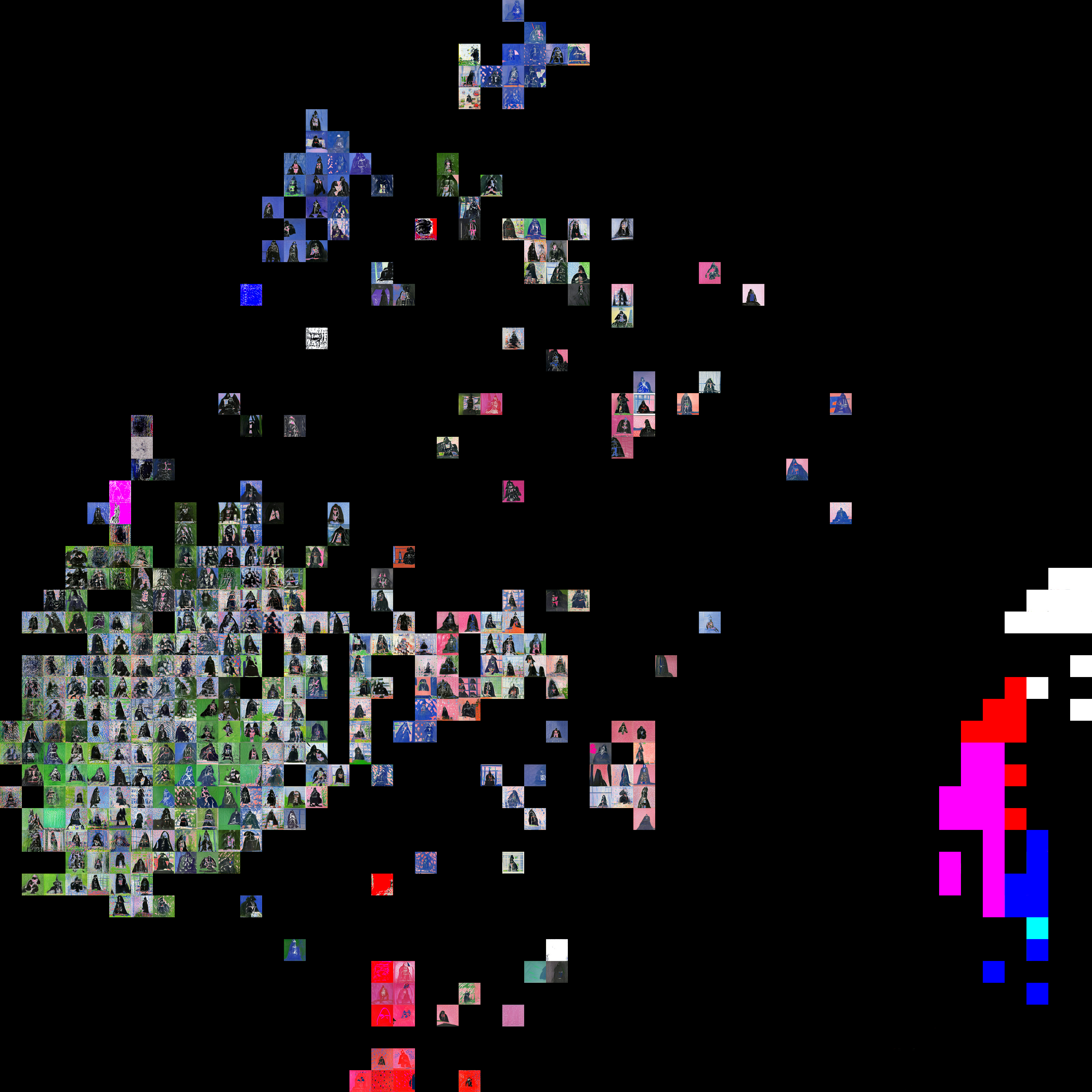}%
		\label{fig:tsne_vader_adam}}
	\hfil
	\subfloat[CMA-ES]{\includegraphics[width=0.32\linewidth]{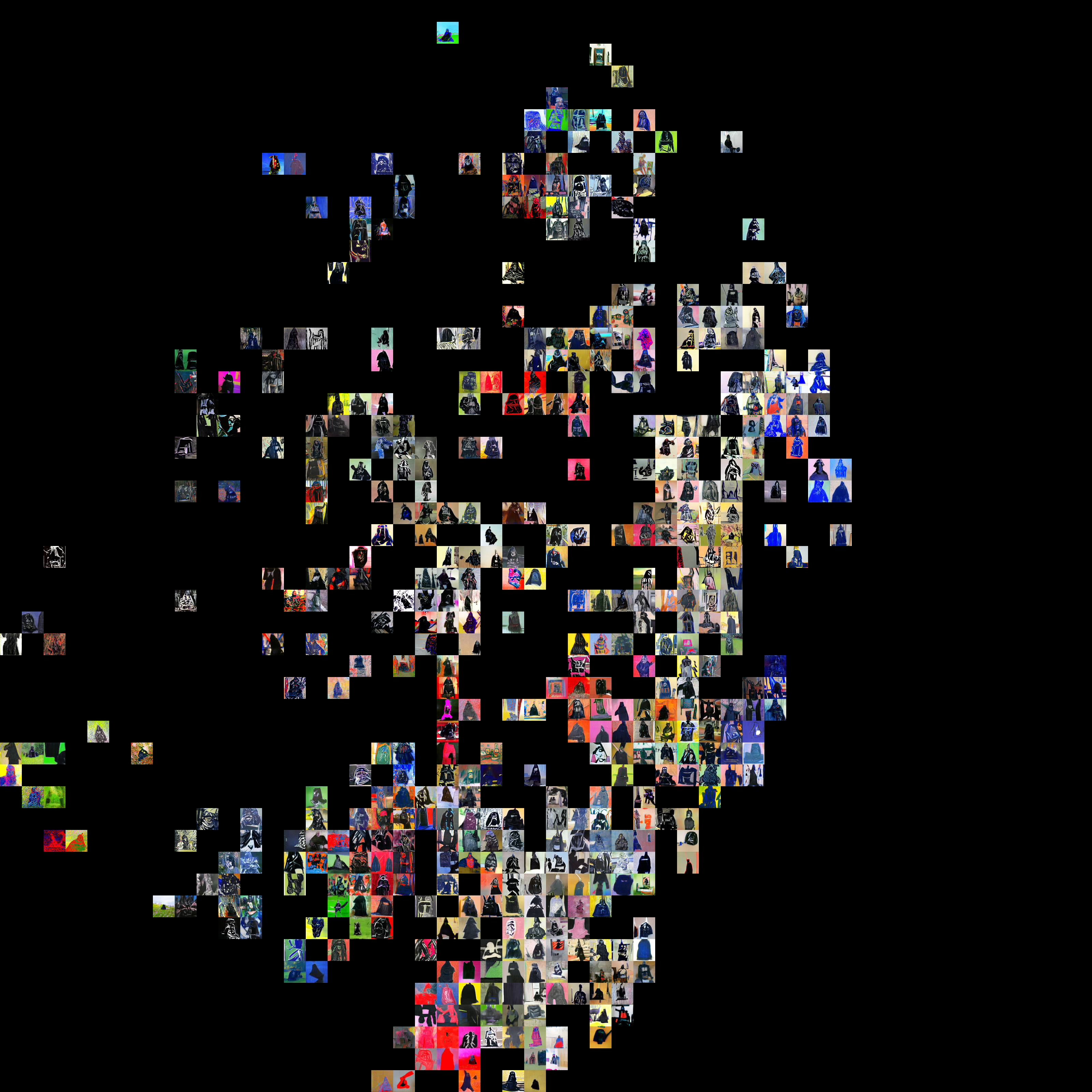}%
		\label{fig:tsne_vader_cmaes}}
	\hfil
	\subfloat[Hybrid]{\includegraphics[width=0.32\linewidth]{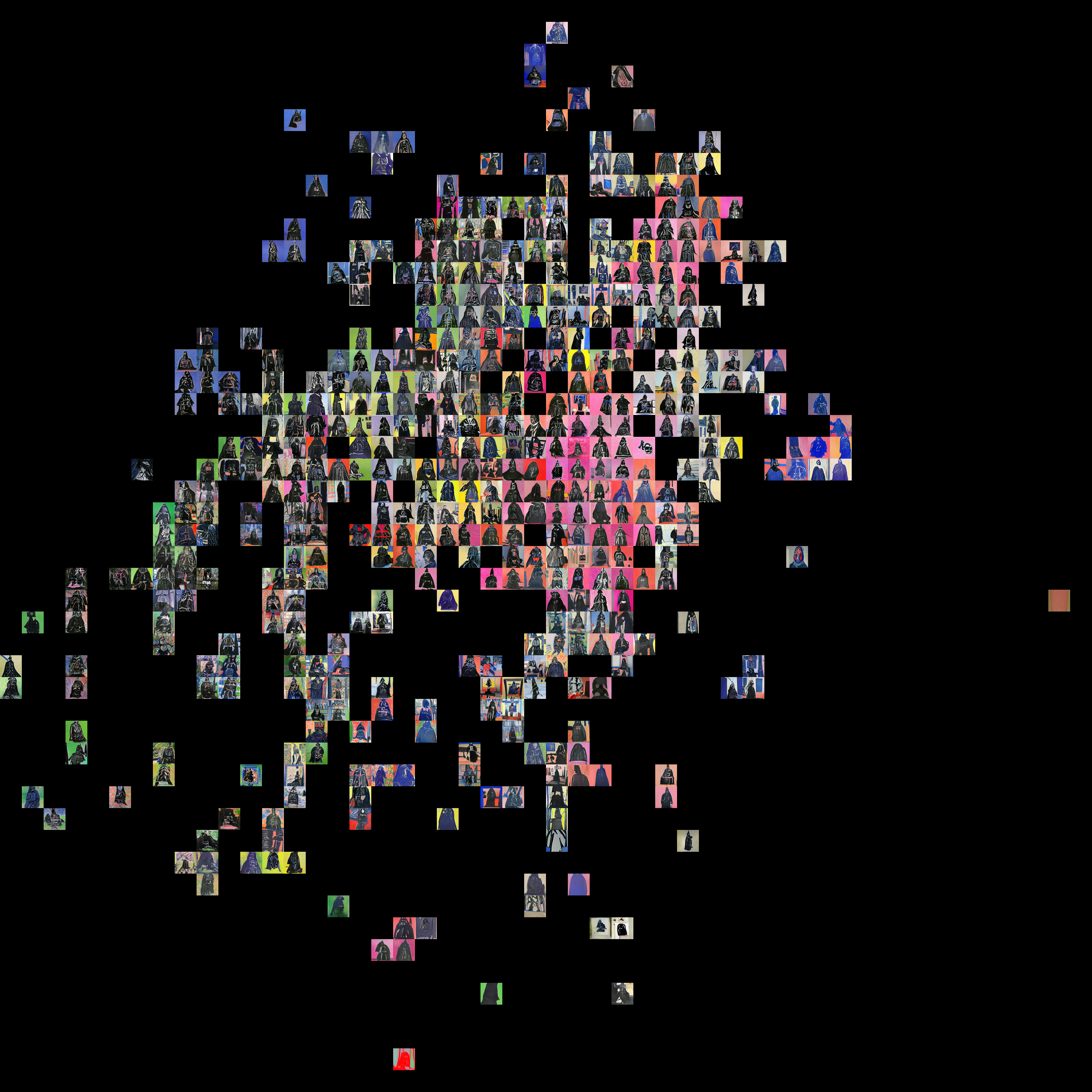}%
		\label{fig:tsne_vader_hybrid}}
	\hfil
	\caption{Two-dimensional grid revealing the distribution of images generated using Adam~\subref{fig:tsne_vader_adam}, CMA-ES~\subref{fig:tsne_vader_cmaes}, and the Hybrid approach~\subref{fig:tsne_vader_hybrid} for the text input "A painting of Darth Vader in the style of Matisse" (images with full resolution available at https://github.com/vfcosta/clip-gan-es/tree/main/images/vader).}
	\label{fig:tsne_vader}
\end{figure*}

Figure~\ref{fig:tsne_vader} shows the distributions for the third text input: "A painting of Darth Vader in the style of Matisse".
The distributions show the same characteristics of the other two text inputs, i.e., the CMA-ES and the Hybrid approaches present more diversity than Adam.
Considering the hybrid approach as the baseline, the results for the Jaccard Index for the Adam and CMA-ES approaches are $0.3015\pm0.0119$ and $0.3671\pm0.0132$, respectively.
Furthermore, we can see that the Adam approach produces a relevant number of samples that contain only a single color without any textures, failing to capture the characteristics of the text input.
This indicates that the evolutionary algorithm also contributes to discarding poor latent vectors, focusing on areas that actually give some significant attributes with respect to the input text.

\begin{figure}[htb]
	\centering
	\subfloat{\includegraphics[width=0.25\linewidth]{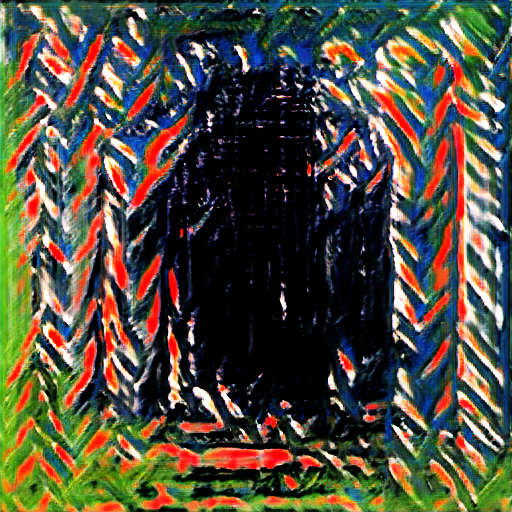}}
	\hfil
	\subfloat{\includegraphics[width=0.25\linewidth]{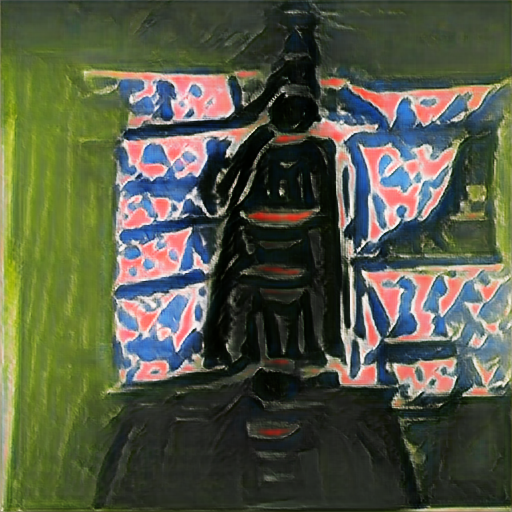}}
	\hfil
	\subfloat{\includegraphics[width=0.25\linewidth]{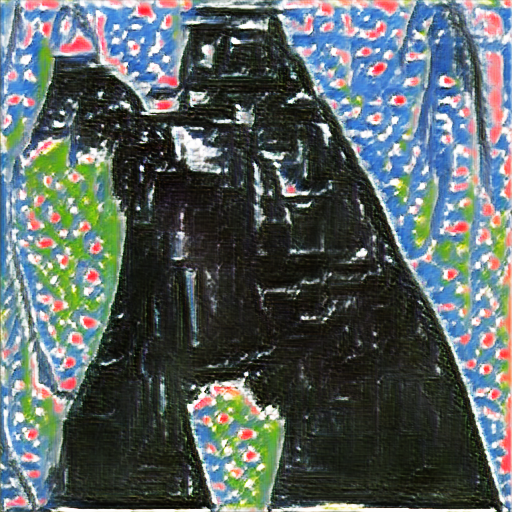}}
	\hfil
	\subfloat{\includegraphics[width=0.25\linewidth]{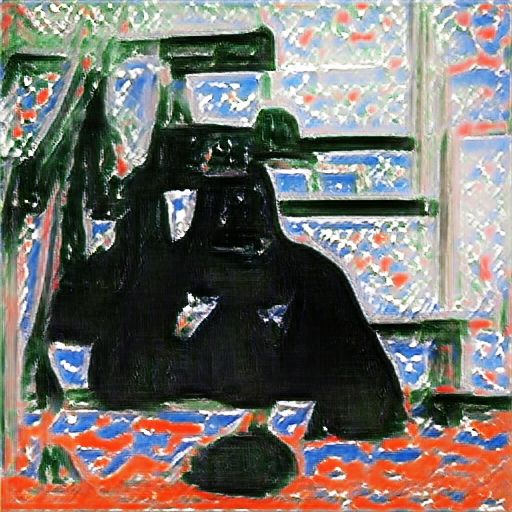}%
		\label{fig:samples_vader_adam}}
	\hfil
	\subfloat{\includegraphics[width=0.25\linewidth]{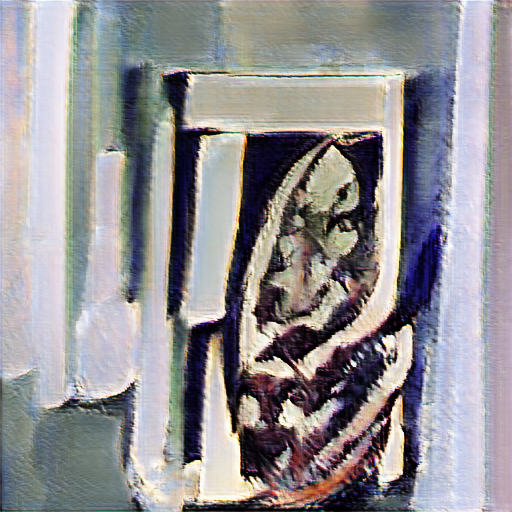}}
	\hfil
	\subfloat{\includegraphics[width=0.25\linewidth]{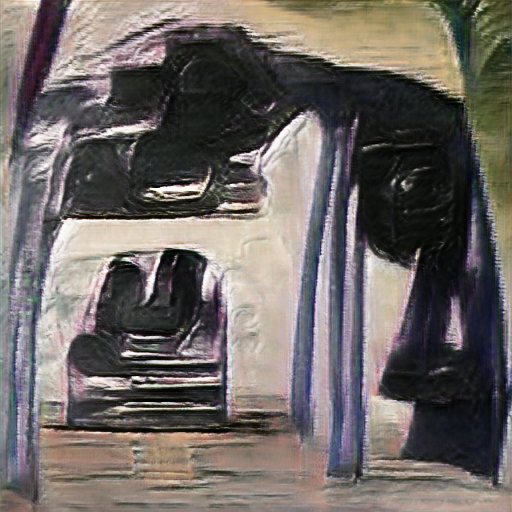}}
	\hfil
	\subfloat{\includegraphics[width=0.25\linewidth]{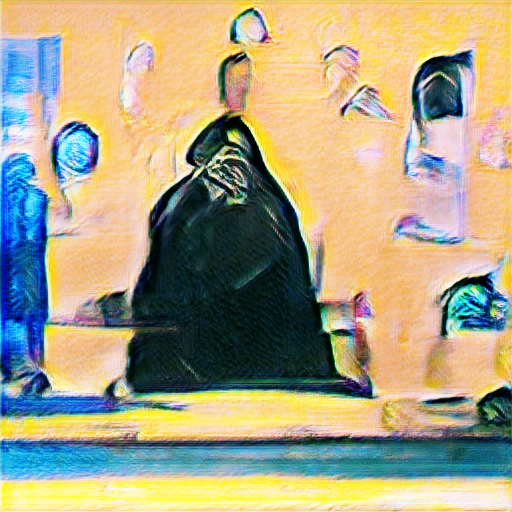}}
	\hfil
	\subfloat{\includegraphics[width=0.25\linewidth]{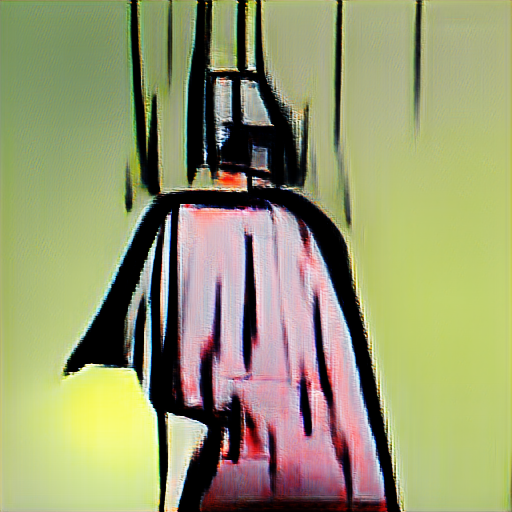}%
		\label{fig:samples_vader_evo}}
	\hfil
	\subfloat{\includegraphics[width=0.25\linewidth]{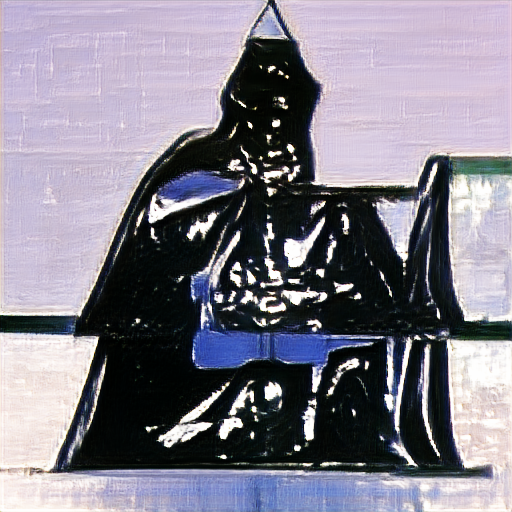}}
	\hfil
	\subfloat{\includegraphics[width=0.25\linewidth]{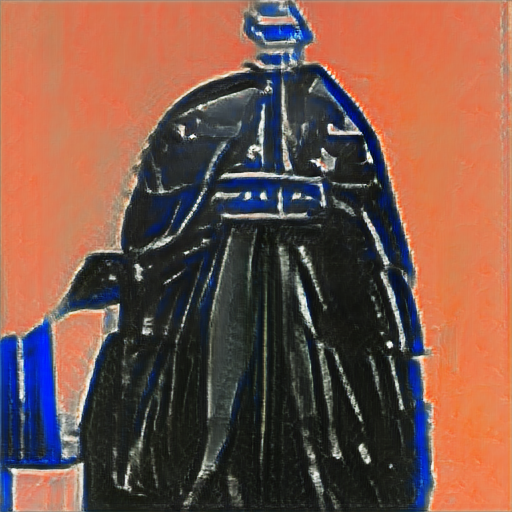}}
	\hfil
	\subfloat{\includegraphics[width=0.25\linewidth]{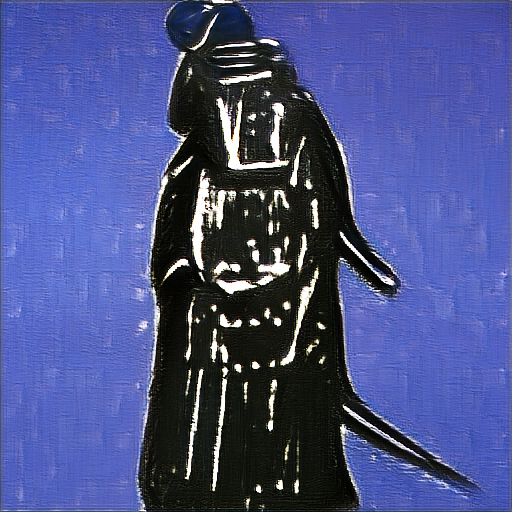}}
	\hfil
	\subfloat{\includegraphics[width=0.25\linewidth]{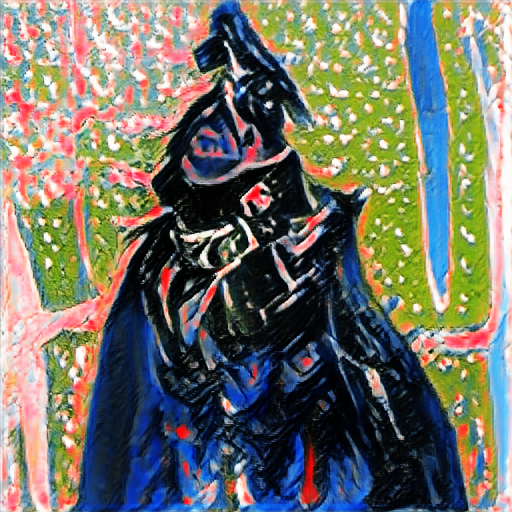}%
		\label{fig:samples_vader_hybrid}}
	\hfil
	
	\caption{Samples from Adam (top row), CMA-ES (middle row), and the Hybrid (bottom row) approach for the text input "A painting of Darth Vader in the style of Matisse".}
	\label{fig:samples_vader}
\end{figure}

We highlight in Figure~\ref{fig:samples_vader} samples achieved from each strategy applied in our experiments.
The top, middle, and bottom rows contain samples generated by applying Adam, CMA-ES, and the Hybrid approach.
We can see in these samples the intense use of colors, as in the paintings of Henri Matisse.
This indicates that the model using CLIP and BigGAN is able to capture the characteristics of the artist mentioned in the input text, transposing his style to the produced samples, and also including the popular fictional character cited in the sentence.

\begin{figure}[bth]
	\includegraphics[width=1\linewidth]{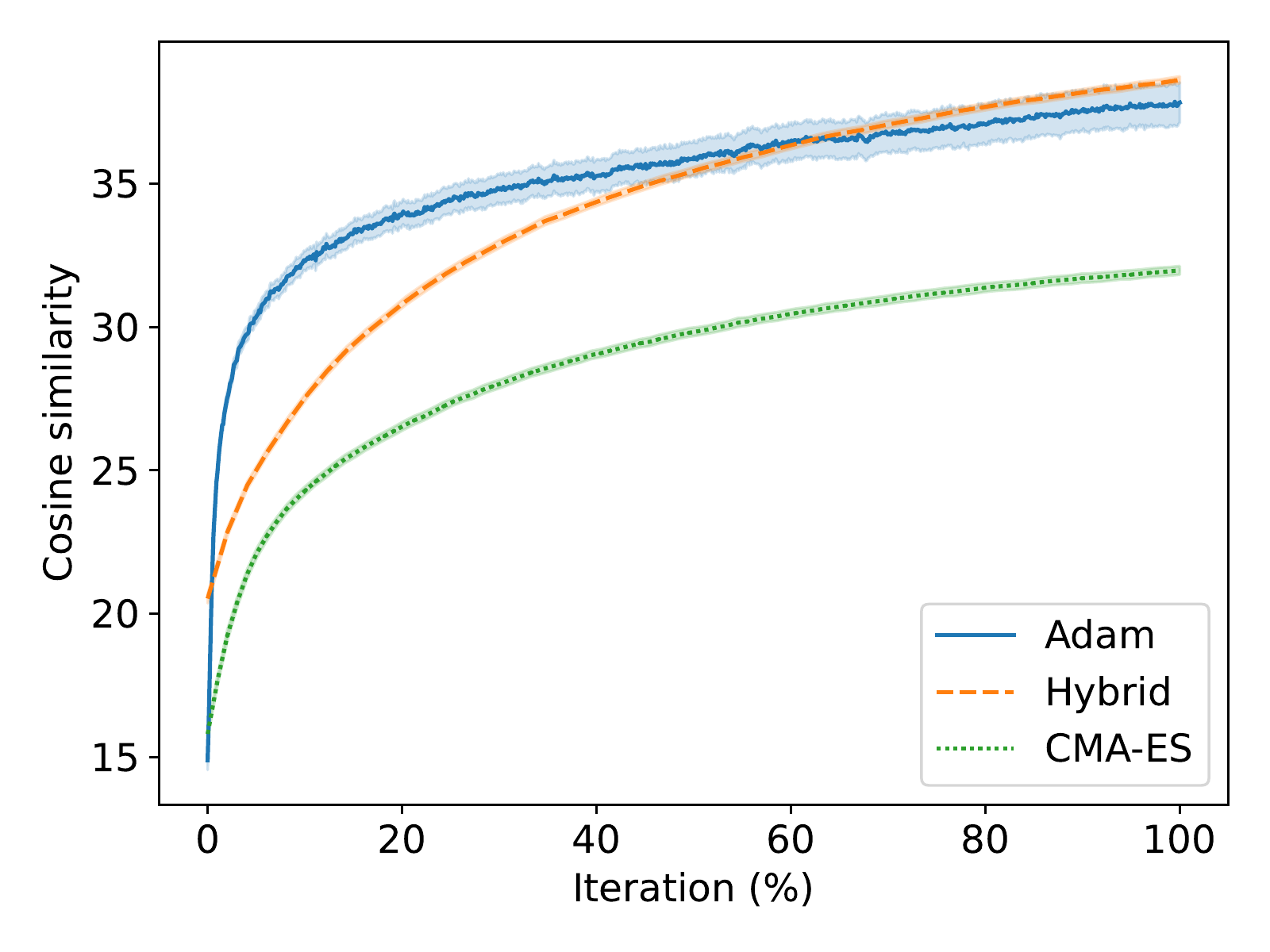}
	\caption{The cosine similarity for best images generated using Adam, CMA-ES, and the Hybrid approach for the text input "A painting of Darth Vader in the style of Matisse".}\label{fig:fitness_vader}
\end{figure}

Figure~\ref{fig:fitness_vader} shows the cosine similarity for the best images generated during the latent space exploration using Adam, CMA-ES, and the Hybrid approach for the text input "A painting of Darth Vader in the style of Matisse".
In this case, the Hybrid approach is able to overcome Adam, achieving better similarity scores.
One possible explanation is that the monochromatic samples that are generated for Adam are degrading the results.

\section{Conclusion}
In this work, we explore the latent space of Generative Adversarial Networks by using Evolution Strategies to improve the diversity of the generated samples.
For this, we adapt a model that uses CLIP and BigGAN for text-to-image generation.
We design an experimental study comparing the exploration of latent spaces by backpropagation with Adam, an evolutionary algorithm using CMA-ES, and a hybrid approach using Adam and CMA-ES.
To evaluate the results, we adapt a method to visualize and quantify the distribution of samples by projecting them into two-dimensional grids.

Our results show that the CMA-ES algorithm contributes to a better exploration of the latent space, achieving better diversity when compared to Adam.
The hybrid leverages the quality and diversity of the samples, outperforming the standalone CMA-ES and Adam approaches.

In future work, we intend to evaluate the results using other generative models such as VQGAN.
Furthermore, we will explore new fitness functions based on the distances of the samples given the distribution maps calculated through the evaluation method.
In this way, we provide an intuitive method to influence the characteristics of the resulting samples.

\bibliographystyle{IEEEtran}
\bibliography{references}

\end{document}